\documentclass{article}

\usepackage{arxiv}

\usepackage{graphicx}
\usepackage{algorithm}
\usepackage{algpseudocode}
\usepackage{amssymb}
\usepackage{amsmath}
\usepackage[table]{xcolor}
\usepackage{caption}
\usepackage{hyperref}
\usepackage{subcaption}
\usepackage{cite}

\newcounter{example}[section]
\newenvironment{example}[1][]{\refstepcounter{example}\par\medskip
   \noindent \textbf{Example~\theexample. #1} \rmfamily}{\medskip}

\title{Coalitional strategies for efficient individual prediction explanation}

\author{Gabriel Ferrettini\\
Université de Toulouse-Capitole,\\
IRIT, (CNRS/UMR 5505)\\
gabriel.ferrettini@irit.fr
\and 
Elodie Escriva\\
Kaduceo\\
elodie.escriva@kaduceo.com
\and
Julien Aligon\\
Université de Toulouse-Capitole,\\
IRIT, (CNRS/UMR 5505)\\
julien.aligon@irit.fr
\and
Jean-Baptiste Excoffier\\
Kaduceo\\
jbe@kaduceo.com
\and
Chantal Soulé-Dupuy\\
Université de Toulouse-Capitole,\\
IRIT, (CNRS/UMR 5505)\\
chantal.soule-dupuy@irit.fr
}

\begin{document}

\maketitle

\begin{abstract}
As Machine Learning (ML) is now widely applied in many domains, in both research and industry, an understanding of what is happening \textit{inside the black box} is becoming a growing demand, especially by non-experts of these models.
Several approaches had thus been developed to provide clear insights of a model prediction for a particular observation but at the cost of long computation time or restrictive hypothesis that does not fully take into account interaction between attributes. This paper provides methods based on the detection of relevant groups of attributes -named \textit{coalitions}- influencing a prediction and compares them with the literature. Our results show that these \textit{coalitional} methods are more efficient than existing ones such as SHapley Additive exPlanation (\textit{SHAP}). Computation time is shortened while preserving an acceptable accuracy of individual prediction explanations. Therefore, this enables wider practical use of explanation methods to increase trust between developed ML models, end-users, and whoever impacted by any decision where these models played a role.

%Machine learning has proven increasingly essential in many fields but a lot of obstacles still hinder its use by non-experts. The lack of trust in the results obtained is foremost among them and has inspired several explanatory approaches in the literature. These approaches provide great insight into the predictions of a model, but at a cost of a long computation time. In this paper, we aim to further improve the detection of relevant attributes influencing a prediction, on the strength of feature selection methods.

\end{abstract}

%\section{TODO}
%\color{red}
%\begin{itemize}
%\item FAIT Reprendre les éléments de SOFSEM nécessaires à une meilleure compréhension du papier ADBIS (notamment introduction de la k-complete)
%\item Compléter l'état de l'art (notamment ce qui a pu sortir de nouveau depuis) et mieux positionner le travail (pourquoi et s'affirmer sur l'orientation sur les coalitions)
%\item FAIT Compléter chaque méthode coalitionnelle avec les figures+commentaires faites dans le mémoire de thèse de Gabriel
%\item FAIT - revoir les formules
%\item FAIT - Compléter les tests de calculs en temps avec la méthode SHAP et tests avec la k-complete. Attention, il faut être cohérent avec les datasets utilisés dans ADBIS (324 datasets provenant d'OpenML)
%\item FAIT - Ajouter les tests pour les alphas + commentaires
%\item FAIT - Ajouter les tests "qualitatifs" sur les explications de prédictions par clustering sur le dataset covid (à détailler également) : même explication pour deux instances différentes et plius gloablement les mesures définies par Elodie (inspirées de la littérature)
%\item Ajouter les tests de comparaisons entre différents type de modèles prédictifs (arbre, randomforest, SVM)
%\item Changer l'intro, la conclusion et le résumé
%\item Changer le titre

%\end{itemize}
%\color{black}
\section{Introduction}

The main deterrent to the comprehension of the majority of machine learning models is their "black box" aspect. Once a  model has been trained, it is often not possible to know the exact reasoning behind the classification performed. Of course, some models remain interpretable by nature, as they are simple enough to be understood when looked at code-wise. As an example, a decision tree is represented as a binary tree and thus will be interpretable for a human unless the number of branches becomes too large to be practical. Some other examples of an interpretable model class are linear methods such as linear, logistic, or Cox regression that are based on a simple linear equation that is easily understandable by a human. The "black box" problem arises when more complex models are used. For those cases, the information necessary to understand directly the model becomes too large to be encompassed for a human. As an extreme example, we can always represent a neural network as a lattice of neurons, but the representation of thousands of nodes and paths would not be useful for a human observer.\par 

Thus, we cannot directly access the information on the internal working of a complex model. Yet it is possible to observe the \textit{effects} of this working, namely, the predictions done by those models. We can consider each prediction made by a model as an additional clue on the way the model functions internally. That the approach is used by a large part of the literature to understand how a model works \cite{adadi2018peeking, carvalho2019machine}. In those approaches, two main goals can be seen: \textit{explaining the model behavior globally}, in order to understand it in the general context, or \textit{explaining the model behavior for a particular prediction}, aiming to highlight that particular decision process.\par

A problem arises when a domain expert user has to study the behavior of particular dataset instances over a predictive model. For example, a physician wanting to perform a cohort study of his patients may also need an explanation of prediction for every single patient, rather than just a global one. In this case, a global explanation is not enough to give the information needed by the study.

To fulfill that need, past researches studied the possibility of explaining single instance prediction of a model, e.g. \cite{strumbelj_efficient_2010} and \cite{casalicchio2018visualizing}. However, these methods may be specifically designed for a single model type as in \cite{lundberg2017consistent}, which limits their use. Some methods are designed to be applicable for all types of models and are more and more used in the industry \footnote{Clinical app that predicts an aggravation risk for a patient hospitalized with Covid-19. Attribute influences are computed with \textit{SHAP}. \url{https://scorecovid.kaduceo.com/}} but suffer from a lack of precision \cite{vstrumbelj2008towards}. Yet overcoming this lack of precision can make their algorithms very long to apply \cite{broeck2020tractability}. This means the field of single instance prediction explanation needs to be developed further, so as to make it usable by everyone.

Our work fits this ambition to help a domain expert user get involved in data analysis operations, especially in learning tasks. Therefore, obtaining explanations for predictive models in an efficient manner, whether in terms of accuracy or computation time, is essential. In a previous work \cite{10.1007/978-3-030-38919-2_26}, we proved the feasibility of lowering the computation time of existing solutions with a limited loss of explanation accuracy.
%This paper proposes methods to find more efficient approximations of these solutions through the exploration of new ways to find groups of attributes.
Finding more efficient approximations of these solutions, thanks to the detection of more relevant coalitions of attributes was the objective of our previous work in \cite{DBLP:conf/adbis/FerrettiniAS20}. This paper extends these proposals by detailing and adding examples for each of the coalitional proposals. A new comparison with one of the most used method of the literature, SHAP \cite{lundberg_unified_2017}, is presented. In particular, our experiments offer a complete view of the performances (in time and error scores) between all the methods as well as a better characterization of the groups of attributes generated. A real use case, regarding the SARS-COV2 data, illustrates the relevance of our coalitional explanations.

The paper is organized as follows. Section \ref{sec:positioning} explores previous works done in the domain of prediction explanation. In particular, the identification of attributes having a significant influence on a model is fundamental. To that end, the automated discovery of groups of linked attributes is an important challenge to overcome. For this purpose, we rely on attribute grouping methods from the literature inspired, notably by feature selection methods.
Then, Section \ref{sec:complete_and_its_approximation} describes the base methods used to generate prediction explanations.
The extension of our work \cite{10.1007/978-3-030-38919-2_26} is proposed in Section \ref{sec:new_methods} to find faster explanation methods. This is achieved through new ways to find groups of attributes for the coalitional method described in Section \ref{sec:SOA_explanation3}.
Experiments are presented in Section \ref{sec:experiments}, showing the interest of our methods, compared to the literature, in terms of computation time and their limited impacts on accuracy loss, significantly improving the results of \cite{10.1007/978-3-030-38919-2_26}.
A particular focus is proposed about the characteristics of the groups of attributes generated by our methods. Then, a qualitative comparison between SHAP and one of our methods are challenged through a real use case and showing the consistency of our approach.
Finally, Section \ref{sec:conclusion} concludes the paper by discussing the perspectives of works including the new possibilities opened by our results.

\section{Related works}
\label{sec:positioning}

\subsection{How to explain a model by using its predictions}
\label{sec:SOA_explanation1}

Explaining the influence of each attribute of a dataset on the output of a predictive model has been explored largely. A few of the works related to global attribute importance on a model can be seen in  \cite{altmann_permutation_2010} \cite{kira1992practical}. The most recent methods are based on swapping the values of attributes in the dataset and analyzing which swaps affect the trained model predictions the most. The more modifying the values of the attributes affect the predictions, the more this attribute is considered important for the model, as a whole. These methods are often used during feature selection, allowing to opt-out attributes useless for the model. 
Another approach for understanding a model is described by Helenius et al. in \cite{henelius2017interpreting}. The authors seek to understand which attributes of the datasets are "linked" to each other, according to the model. In particular, they proceed by randomizing the values of potential groups of attributes: when the predictions vary more than a predetermined threshold, the attributes are linked together. %This process creates a grouping of the dataset attributes, which inform the users on how they are interacting with each other according to the model.
\par

The problem with a fixed global influence is that predictive models are often not consistent with the whole dataset on which they are trained. These global influences give an insight into the general working of the studied model, but there will often be particular regions of the dataset where the model deviates from this general influence. Thus, there also exists a need for a single instance prediction explanation, showing the user how a particular instance is classified, independently to the rest of the dataset. These methods aim to provide insight into the global influence of each attribute, rather than on their influence on a single prediction.\\

In the field of single prediction, these global influence methods have served as a basis to \cite{casalicchio2018visualizing}, using the same randomization technique. Given a single instance, the importance of each attribute is obtained by looking at the evolution of the prediction performance of the model on the instance when all of its values are swapped with other values of the dataset except the value of the attribute being studied. The more the prediction varies with the values swapping, the less the fixed attribute is important for the prediction. This method has the interest of relying on the same principle as the global technique which is largely recognized, but the main caveat is its computational cost needed for explaining the prediction of only one instance, as a large number of new predictions has to be generated for each single instance prediction to explain. Another caveat is this prediction explanation is realized from the point of view of model performance. Meaning that their metric shows which feature improves the performance of the model, rather than which feature the model consider as important for its prediction. If this line of reasoning is interesting for the model explanation field, it does not correspond to our scope, as we are aiming to help users understand how a model works, and not how to improve it.\par

Another approach can be found in \cite{wachter2017counterfactual}, which inspired many of the functions of \cite{wexler2019if}. The principle here is to determine the smallest change needed in order to change the classification of an instance by the model. It has been integrated into the What-if tool. This tool allows the user to select a particular data point in the dataset, and visualize the nearest point classified differently. This is displayed along with the differences between the two points, thus highlighting what let the two points to be classified differently. \par 

%\begin{example}
%This method is often used when studying and training neural networks, notably in the context of adversarial networks. Adversarial networks are two neural networks trained in opposition. The first one is trained to perform the wanted task, as classifying an image, while the other is trained to create instances designed to fool the first network. In the example of image classification, the second network will perform a modification as little as possible in order to change the classification of the object. This modification is often not perceivable by the human eye but can change completely the nature of the recognized object. These approaches led to the development of objects like the ones depicted in figure \ref{fig:SOA_adv_glasses}, which are glasses modifying enough the face of the wearer to change completely the person being detected.
%\end{example}

These methods give us an insight into how the model works, as they display the attributes which could put the instance in another class if their value was changed slightly. However, if these methods are interesting for analyzing the important points of a model, this kind of information would be far less useful to a domain expert, as they already require the user to understand the importance of this information and draw conclusions on it by himself.

%\begin{figure}[htbp]
%    \centering
%    \includegraphics[width = \textwidth]{Extension_ADBIS_ISF/images/whatif_counterfactual.png}
%    \caption{The What-if tool allows the user to see the nearest point in the dataset classified differently than the one selected. On the left is highlighted the main differences of the two points, which led them to be classified differently.}
%    \label{fig:whatif_counterfactual}
%\end{figure}

%\begin{figure}[htbp]
%    \centering
%    \includegraphics[width = \textwidth]{Extension_ADBIS_ISF/images/adversarial_glasses.png}
%    \caption{Adversarial glasses which fools facial recognition neural networks into classifying the wearer (top image) as another person (bottom image)}
%    \label{fig:SOA_adv_glasses}
%\end{figure}

One of the early explanation methods, found in \cite{vstrumbelj2008towards}, avoid the drawbacks mentioned above. In this explanation method, the weight of an attribute, on a particular prediction, is estimated by the difference of influence over the model with and without this attribute. This absence of an attribute is simulated by a weighted mean of the predictions of the model with all the possible values of the attribute, weighted by their probability of appearing in the dataset. This is faster than the method of \cite{casalicchio2018visualizing} as only one value is randomized. Later, in \cite{strumbelj_efficient_2010}, the possibility of retraining the model entirely without the considered attribute in the dataset is proposed, which consists of an interesting trade-off: the time needed for retraining a model for each attribute of a dataset can be considerable, but once this training has been done, the prediction comparison can become near-instantaneous. These methods are named \textit{SHapley Additive exPlanations} by Lundberg et al. \cite{lundberg_unified_2017}, who regrouped a large number of similar methods of explanation, and detailed in Section \ref{sec:complete_and_its_approximation}.

\cite{lundberg_unified_2017} highlights several interesting properties about these methods :
\begin{itemize}
    \item Local precision: The system describes precisely the model in the close vicinity of the explained instance.
    \item "Missingness": If an attribute is missing for the prediction, the method does not give it a weight, or gives it a weight of zero.
    \item Consistency: If the explained model changes in a way that makes an attribute more important, or does not change its importance, its attributed weight is not diminished. This property is important, as some of the early prediction explanation methods could have an erratic behavior in some cases, as shown in an example of \cite{lundberg_unified_2017}.
\end{itemize}

This type of prediction explanation is quite interesting, as we are aiming to facilitate the understanding of any machine learning models for users without particular knowledge on data analysis or machine learning. Thus, it is more relevant to focus on the works as \cite{Strumbelj:2010:EEI:1756006.1756007} or \cite{datta_algorithmic_2016}, cited as \textit{additive} methods, as they generate a simple set of importance weights for each attribute. This set of weights is easy to interpret, even for someone without expertise in machine learning. Yet, these methods have a major deterrent: their complexity makes them difficult to use for the average user. That is why \cite{lundberg_unified_2017} explored methods to generate explanations faster, but at the cost of very restricting hypotheses, as the independence of each attribute of the dataset, or the linearity of the model, which is not always the case.

%A vérifier :
\textit{LIME} is another popular additive interpretability method \cite{ribeiro2016should}. However, it suffers from previously described restricting hypotheses as it relies on a surrogate linear model to locally approximate influence of each variable, through multiple random perturbations of the instance to explain. This random nature of \textit{LIME} leads to an high instability, especially compared to \textit{SHAP} \cite{elshawi2020interpretability}. Thus we will focus in this article on Shapley-based interpretability methods.

The ability to explain the prediction of any model thus appears to be a key point for allowing a broader public (non-expert) to access and use machine learning models.

This need led us to consider the diverse explanation systems, developed in the literature, as having a major interest in giving more autonomy to domain experts performing data analysis tasks. Yet, the computational load found in the most generic methods can be a hindrance to their use. In this paper, we seek to propose a prediction explanation method as generic as possible and try lowering its computing time without losing too much information.\par

\subsection{Grouping attributes}
\label{sec:SOA_explanation3}

In our work, we want to facilitate the generation of prediction explanation, without having to restrict to a given set of models. This paper is the continuity of \cite{10.1007/978-3-030-38919-2_26} in which we already established possible methods of simplification and rely on the automatic detection of groups of attributes (especially the \textit{K-complete} method, detailed in Section \ref{sec:k-complete}.).

For this work, we aim to identify and compare additional methods detecting groups, in order to compare their influence on the efficiency of the simplification method.

The selection of relevant attributes to be grouped can take inspiration from the works in the field of feature selection \cite{Bolon2013} \cite{Yu2004}. In particular, the methods proposed in a dimensional reduction goal seem to reach our scope. Indeed, these methods have to automatically detect interactions between attributes for reducing a potential high dimensionality in a dataset. Thus, two main approaches, feature extraction (mainly the principal component analysis) and filter methods (which measure the relevance of features by their correlations) can be considered.
The fact that the principal component analysis (PCA) and the filter methods rely only on information provided by a dataset (independent of the model used in analysis) is a great advantage for our work, in contrast with techniques such as SVM-RFE \cite{Rakotomamonjy2003} or FS-P \cite{Mejia2006}, based on a specific model.
Indeed, different predictive models can classify differently the same instance. Thus, an explanation on this instance can be different from one model to another, and cannot depend on a selection of influence attributes made by a unique predictive model, such as SVM.
The PCA is a largely recognized method to provide new features from sets of correlated attributes. 
The Correlation-based feature selection (CFS) methods \cite{hall1999correlationbased} are promising candidates. In particular, the use of a multicollinearity measure by a variance inflation factor (VIF), can provide sets of attributes having linear correlations between them. This avoids calculating collinearity between pairs of attributes, using Pearson's measure, for example.
However, the VIF measure is unable to compute non-linear correlations, on the contrary of the Spearman correlation factor. Even if this factor only works between pairs of attributes, the capacity to detect non-linear correlations makes it a good candidate.

\section{The $Complete$ method and its approximations from the literature}
\label{sec:complete_and_its_approximation}

This section introduces the $Complete$ method considered as the baseline of our work (\cite{10.1007/978-3-030-38919-2_26}), computing all possible sub-groups of attribute influences. Then, two approximations of the literature, \textit{K-complete} \cite{10.1007/978-3-030-38919-2_26} and \textit{SHAP} \cite{lundberg_unified_2017} are detailed. Their limits are finally discussed, opening a way for new proposals of coalitional methods in Section \ref{sec:new_methods}.

\subsection{Complete explanation}
\label{sec:complete}

To answer the problems of interaction between attributes, we propose to take inspiration from the work of \cite{Strumbelj:2010:EEI:1756006.1756007}. The prediction task is a framework close to the situation called "coalitions", where groups of attributes can influence the prediction of the model. In this context, each attribute cannot be considered as independent, but in all possible attributes combinations. The influence of an attribute is measured according to its importance in each coalition. We can then refer to the coalition games as defined by Shapley in \cite{shapley1953}: A coalitional game of $N$ players is defined as a function mapping subsets of players to gains $g: 2^{N} \mapsto \mathbb{R}$. The parallel can easily be drawn with our situation, where we wish to assess the influence of a given attribute \textit{in every possible coalition of attributes}. We then look at not only the influence of the attribute but also its use in all subsets of attributes. We thus define the \textit{complete influence}  of an attribute $a_{i} \in A$ on the classification of an instance $x$ : given a dataset of instances described along the attributes of $A$, the \textit{complete} influence of the attribute $a_{i}$ on the classification of an instance $x$ by the classifier confidence function $f$ on the class $C$ is dependant on the influence of all the possibles subgroups $A' \subseteq A$ which does not contain $a_i$. Thus, the \textit{complete} influence of $a_i$ is :

\begin{equation}
\mathcal{I}^{C}_{a_{i}}(x) = \sum_{A' \subseteq A \backslash a_{i}} p(A',A)*(inf^{C}_{f,({A'\cup a_{i}})}(x) - inf^{C}_{f,A'}(x))
\label{complete_equation}
\end{equation}

With $p(A',A)$ a penalty function accounting for the size of the subset $A'$. 
Indeed, if an attribute changes a lot the result of a classifier, in a large group of attributes, it can be considered as very influential compared to the others. On the opposite, an attribute changing the result of a classifier, whereas this classifier is based on a few attributes, cannot be considered to have a decisive influence.
The Shapley value \cite{shapley1953} is a promising candidate and defines this penalty as:
\begin{equation}
p(A',A) = \frac{|A'|!*(|A|-|A'|-1)!}{|A|!}
\end{equation}

This \textit{complete influence} of an attribute now takes into consideration its importance among all the possible attribute configurations, which is closer to the original intuition behind attributes' influence. However, computing the \textit{complete influence} of a single instance is extremely computationally expensive, with complexity in $\bigcirc(2^n*l(n,x))$, with $n$ the number of attributes, $x$ the number of instances in the dataset, and $l(n,x)$ the complexity of training the model to be explained.
It is then not practical to use the \textit{complete influence}. Consequently, it becomes necessary to seek a more efficient way to explain predictions. Although the \textit{complete influence} is too computationally heavy, it can be considered as an excellent baseline \cite{Strumbelj:2010:EEI:1756006.1756007}. Thus, we can evaluate other explanation methods by studying their differences with the \textit{complete influence}.

\begin{figure}[h]
    \centering
    \includegraphics[width = \linewidth]{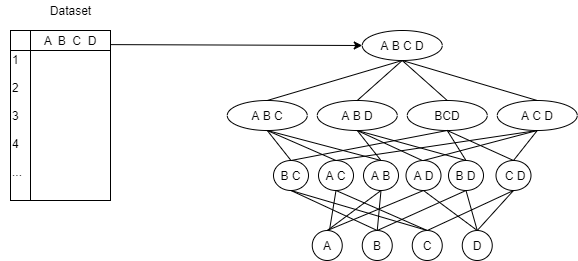}
    \caption{Depiction of the groups calculated by the complete method for a dataset with 4 attributes. Each possible combination of attributes is calculated to ensure an influence value as close to the reality as possible.}
    \label{fig:2_complete}
\end{figure}

\begin{example}
As depicted in Figure \ref{fig:2_complete}, the influence of an attribute depends on its influence alone, but also on each possible groups of attributes containing it. As in the Figure \ref{fig:2_complete}, for a dataset with 4 attributes $A~B~C~D$, the influence of the attribute $A$ is composed of the influence of $\{A\}$ alone, along with the influences of the groups $\{A,B\}$, $\{A,C\}$, $\{A,D\}$, $\{A,B,C\}$, $\{A,B,D\}$, $\{A,C,D\}$ and $\{A,B,C,D\}$.
\end{example}

\subsection{K-complete explanation}
\label{sec:k-complete}

%TODO: A EXPLIQUER For this we cannot fully rely on recent works (e.g. \cite{Strumbelj:2010:EEI:1756006.1756007} and \cite{lundberg_unified_2017}), as explained in Section \ref{chap:SOA_reco}. 
%An approximation of the \textit{complete influence} has to remain accurate and practical, as much as possible. For this we cannot fully rely on recent works (e.g. \cite{Strumbelj:2010:EEI:1756006.1756007} and \cite{lundberg_unified_2017}), as explained in Section \ref{chap:SOA_reco}. 

Another possible approach is proposed in our previous work \cite{10.1007/978-3-030-38919-2_26}.
This approximation, looking for a subset of all the subgroups of the $complete$ method, could be more practical in terms of complexity. This solution should produce explanation, a priori, more accurate than the basic consideration of independent attributes (\textit{linear influence}). We consider then the \textit{depth-$k$ complete influence} defined as the complete influence, but ignoring the groups of attributes $A'$ with a size superior to $k$ :
\begin{equation}
\mathcal{I}^{C_{k}}_{a_{i}}(x) = \sum_{A' \subseteq A \backslash a_{i},~ {|A'| < k}} p_k(A',A)*(inf^{C}_{f,({A'\cup a_{i}})}(x) - inf^{C}_{f,A'}(x))
\label{depth_k}
\end{equation}

\begin{equation}
p_k(A',A) = \frac{|A'|!*(|A|-|A'|-1)!}{k*(|A|-1)!}
\label{shap_k}
\end{equation}

In particular, we can note that the \textit{linear} influence is actually identical to the \textit{depth-$1$ complete influence}. 
The intuition behind this approach is to eliminate the larger groups, which have a lesser impact on the Shapley value while being the most costly to calculate.

\begin{figure}[h]
    \centering
    \includegraphics[width = \linewidth]{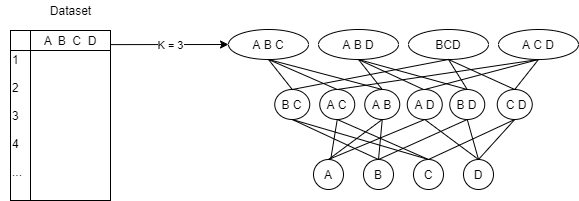}
    \caption{Depiction of the groups calculated by the k-complete method for a 4 attributes dataset. The group size is limited by the parameter k: here the groups maximum size is 3.}
    \label{fig:2_k-complete}
\end{figure}

\begin{example}
As depicted in \ref{fig:2_k-complete}, for the same dataset with 4 attributes and a parameter $k=3$, the total influence of the attribute $A$ only depends on the influence of $A$ alone and the groups of attributes containing $A$ and with a maximum size of $3$ : $\{A,B\}$, $\{A,C\}$, $\{A,D\}$, $\{A,B,C\}$, $\{A,B,D\}$ and $\{A,C,D\}$.
\end{example}

\subsection{SHapley Additive exPlanation}

As explained in Section \ref{sec:SOA_explanation1}, a major contribution in the literature about single prediction explanation can be found in \cite{lundberg_unified_2017}.
This approach is one of the most used in the context of prediction explanation, especially in the biology and medical fields like in \cite{10.1093/nar/gkaa219} and \cite{9233366, Bibault2020, lauritsen2020explainable}.
\cite{lundberg_unified_2017} theorizes a category of explanation methods, named \textit{SHapley Additive exPlanations} methods, and produces an interesting review of the different methods developed in this category. The authors first propose to overcome a growing problem caused by various proposals of explanation methods: when is it preferable to use one of these methods rather than another one?

From this observation, the paper describes a unified approach for 6 existing methods \cite{Ribeiro2016}, \cite{shrikumar_learning_2017}, \cite{10.1371/journal.pone.0130140}, \cite{Lipovetsk2001}, \cite{Strumbelj:2010:EEI:1756006.1756007} and \cite{datta_algorithmic_2016}.
They are summarized in \cite{lundberg_unified_2017} as methods giving for a particular prediction a weight to each attribute of the dataset.\par
This creates a very simple "predictive model", locally mimicking the original model's behavior. Thus, we have a simple interpretable linear model that gives information on the original model's inner working in a small vicinity of the predicted instance. The methods, from which these weights are affected to each attribute, vary between the different \textit{additive} methods, but the end result is always this vector of weights.\par
In particular, this unified approach is based on a \textit{SHAP} (SHapley Additive exPlanation) value giving the importance of a feature depending on the predictive model used (for a specific prediction):  \textit{LinearSHAP} for linear models, \textit{TreeSHAP} for tree-based models and \textit{DeepSHAP} for neural networks. A \textit{KernekSHAP} is also presented and can be used for any type of predictive model but at the cost of a longer computation time.

%For instance, \cite{elshawi2019ilime} proposes an intuitive method for explaining the prediction of a model on a single instance. First, new random data points are generated in the close vicinity of the studied prediction. The analyzed model is then asked to perform predictions for those points. Hence, a "sub dataset", comprised of those new points and their affixed predictions. A linear model is then trained on this sub dataset: thus, the linear model mimics the behavior of the original model in the vicinity of the explained data point. As a linear model is simply a set of weights and conditions on each attribute of the dataset, it is interpretable for a human, and can be used as a description of the original model's behavior for the prediction on the data point.\par
%Another method is the one described in \cite{Strumbelj:2010:EEI:1756006.1756007}. In this article, they describe a method that relies on comparing the differences between the predictions of the model in different configurations. In this method, the importance of an attribute is considered as the difference in the prediction of the model when the attribute is added to the dataset. Thus, the weight assigned to an attribute will be dependent on the information brought by this attribute to the prediction.

Despite this interesting strategy, this approach is very time-consuming, as showed in \cite{broeck2020tractability}.

The full implementation of this work can be found here: \url{https://github.com/slundberg/shap}

\subsection{Limitations of these approaches}

All of the three approaches described in this section have several limitations. \textit{Complete} method has an exponential complexity with respect to the attribute number which makes it unusable in most practical cases. 
It can be approximated by \textit{SHAP} methods -\textit{KernelSHAP} and its variants- but at the cost of very restrictive hypothesis such as local linearity that does not fully account of dependence between attributes, thus biasing the explanation results. Moreover, the computation may take a very long time in such a configuration of high attribute interdependence, which is often the case in practice.
The \textit{k-complete} methods is another way to approximate the \textit{complete} one that gives the ability to select complexity with the \textit{k} parameter. The inconvenience of this set of methods is that these generated groups may include useless or redundant subgroups that greatly increases computation time without any significant gain in accuracy for the \textit{complete}.

\section{Proposals of coalition computing methods}
\label{sec:new_methods}

Limitations of current approximation methods of the \textit{complete} highlighted in the previous section indicate that potential interactions between attributes must be better taken into account.
Combination of unrelated attributes should be avoided at maximum to minimize the complexity, thus computation time, while staying at high accuracy with respect to the \textit{complete} method.
In this end, we propose several grouping methods such as an existing algorithm from \cite{Henelius1175294}, and new algorithms based on \textit{Principal Component Analysis} (PCA), \textit{Spearman correlation factor} (Spearman) and \textit{Variance Inflation Factor} (VIF). We also develop \textit{Reverse} methods -based on either Spearman or VIF- that only gather uncorrelated attributes, since groups only formed of highly correlated attributes contain mostly redundant information.
For each algorithm, a parameter controls the size of the generated subgroups. A higher value of this parameter generates larger groups whereas a smaller value produces smaller, thus less complex, groups.
Explanations through influence for each attribute of the dataset is then computed using \textit{coalitional} influence, which takes as parameter the list of groups generated by a grouping method.

%As we have seen, the coalitional influence seems to be a good candidate, but its reliance on a grouping algorithm to function calls for a deeper exploration of the performances of the \textit{coalitionnal} influence, depending on the grouping algorithm used. We based our first algorithm on the work of \cite{Henelius1175294}. The new algorithms we want to test are based on the Variance Inflation Factor (VIF), the Spearman correlation factor, and the Principal Component Analysis (PCA) of a dataset. For each algorithm, we implement a parameter that controls the size of the subgroups that are generated. A higher value of this parameter generates larger groups whereas a smaller value produces smaller groups.

\subsection{Coalitional explanation methods}

The coalitional explanations, presented in this Section \ref{sec:new_methods}, identify the attributes having an interaction between them. We can obtain a grouping such as $G = \{\{a_1,a_3\}, \{a_2,a_5,a_8\}, \{a_4\}...\}$. With such groupings of attributes, it becomes possible to consider only the attributes of a subgroup, without having to consider every possible attribute combination. It is important to note that the groups do not necessarily have to be exclusive, which mean an attribute $a_i$ can be found in multiples groups of $G$. We then obtain a \textit{coalitional influence} of an attribute $a_i$ : Given $G_{a_i}$, the subset of $G$ containing all the attributes groups $g \in G$ such as $a_i \in g$ 
\begin{equation}
simple\mathcal{I}^{C}_{a_{i}}(x) = \sum_{g' \subseteq g \backslash a_{i},~ g\in G_{a_i}} p_c(g',g,G_{a_i})*(inf^{C}_{f,({g'\cup a_{i}})}(x) - inf^{C}_{f,g'}(x))
\label{coalitionnal_equation}
\end{equation}

\begin{equation}
p_c(g', g, G_{a_i}) = \frac{|g'|!*(|g|-|g'|-1)!}{\sum_{g\in G_{a_i}} |g|!}
\label{shap_c}
\end{equation}
Given the fact that we can set a maximum cardinal $c$ for our subgroups, the complexity is now, in the worst case, $O(2^c*\frac{n}{c}*l(n,x)) \approx O(n*l(n,x))$. This method calculates less groups than the \textit{depth-$k$ complete influence}, but tries to make up for it by only grouping the attributes actually related to each other. In order to determine which attributes seem to be related, several types of coalition strategies are proposed below.

\subsubsection{Model-based coalition}

Regarding the \textit{Model-based coalition} approach, the groups of attributes are created by using the model itself to detect interacting attributes. In this approach, no correlation is detected, but only interaction in the sense of the model usage of the attributes.
This is done by randomizing the values of the dataset and studying the evolution of the model predictions.
It consists of measuring the differences of predictions on the whole dataset before and after the randomization.
When attributes are considered to be part of the same group, their values are swapped together with the values of another instance, classified by the model as the same class as the starting instance. Each attribute outside of the group has its value swapped completely randomly. Once this has been done, the new instances are classified by the model. The ratio of differences between the old and the new classification is called fidelity. A higher fidelity meaning a lower variation of the predictions.
At each iteration, the attribute which removal lowers the less the fidelity is removed until it is not possible to keep the fidelity above a fixed threshold. Then the group is considered as fixed.
This attribute grouping algorithm has been developed in \cite{Henelius1175294} and is detailed in Algorithm \ref{alg:model}.

\begin{algorithm}[htbp]
\begin{algorithmic}
\Require Sensitivity parameter $\delta > 0$, the number of attributes $m$, and a fidelity function $fid()$.
Two auxiliary functions $L(X) = \bigcup_{i \in X}\{\{i\}\}$ and $F(X) = L(\bigcup_{Y \in X} Y)$, which produces sets of singletons (e.g. $L(\{1,2,3\}) = F(\{\{1,2\},\{3\}\} = \{\{1\},\{2\},\{3\}\}$)
\Ensure $\sigma$ a coalition of attributes
    \State $\sigma \leftarrow \{\}$  %\Comment{initialisation of the coalition}
    \State $R \leftarrow \{m\}$ \Comment{R contains a group to test for}
    \State $A \leftarrow \{\}$ \Comment{A contains the removed attributes}
    \State $\Delta \leftarrow fid(L([m]))+\delta$
    \While{$R \ne \{\} or A \ne \{\}$}
    
        \If{$A = \{\}$ and $fid(\{R\} \bigcup F(\sigma)) < \Delta $}
            \State \Comment{if we are already below $\Delta$ before removing any attribute assign the remaining attributes to singleton groups}
            \State $\sigma \leftarrow \sigma \bigcup L(R)$
            \State $R \leftarrow \{\}$
            \State $A \leftarrow \{\}$
        \Else
            \State \Comment{Find an attribute j whose removal from R decreases the fidelity least}
            \State $j \leftarrow argmax_{j \in R} fid(\{\{R\backslash\{j\}\} \bigcup \{\{j\}\} \bigcup \{A\} \bigcup F(\sigma))$
            \If{$|R| = 1$ or $fid(\{\{R\backslash\{j\}\} \bigcup \{\{j\}\} \bigcup \{A\} \bigcup F(\sigma))$}
                \State \Comment{If the fidelity drops below $\Delta$ add the group of attributes to the results and look for the next group of attributes}
                \State $\sigma \leftarrow \sigma \bigcup \{R\}$
                \State $R \leftarrow A$
                \State $A \leftarrow \{\}$
            \Else
                \State \Comment{If the fidelity stays above $\Delta$ continue removing the grouping $R$}
                \State $R \leftarrow R \backslash \{j\}$
                \State $A \leftarrow A \bigcup \{j\}$
            \EndIf
        \EndIf
    \EndWhile
    \State \Return $\sigma$
\end{algorithmic}
\caption{Model-based coalition extraction.}
\label{alg:model}
\end{algorithm}

\begin{figure}[h]
    \centering
    \includegraphics[width = \linewidth]{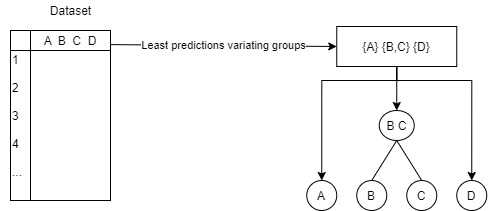}
    \caption{Depiction of the groups calculated by the model-based coalition method for a 4 attributes dataset.}
    \label{fig:coal-model}
\end{figure}

\begin{example}
Given the same dataset with 4 attributes, we apply the algorithm from \cite{Henelius1175294} on the dataset. It aims to build groups as small as possible such as when:
\begin{itemize}
    \item The values of the grouped attributes are randomized \textbf{inside} of their original instance class ;
    \item The values of the non-grouped attributes are randomized completely ;
    \item This randomization is applied to the whole dataset.
\end{itemize}
The predictions of the model for the whole dataset do not vary more than a threshold percentage of $\delta$.\par 
In the first iteration, the algorithm finds that the smallest group of attributes that makes the predictions varies less than the threshold $\delta$ is $\{B, C\}$. Removing $C$ or $B$ makes the predictions vary more than the threshold and as such, the algorithm stores $\{B, C\}$ as a first group.\par
Now the algorithm tries to build another such group with the remaining non grouped attributes and find that $\{A, D\}$ makes the predictions vary too much. Since the biggest remaining possible group is already making the predictions vary more than the threshold, all the non grouped attributes are considered as singletons, resulting in the grouping $\{\{A\},\{B, C\},\{D\}\}$.\par
This grouping is then used to determine how each attribute has its influence calculated. As an example, the total influence of $A$ only consists of the influence of the singleton $\{A\}$, while the total influence of $B$ is composed of the influences of $\{B\}$ and $\{B, C\}$. Similarly, the total influence of $C$ includes the influences of $\{C\}$ and $\{B, C\}$, and as $D$ is in a singleton, its influence only takes into account the influence of $\{D\}$. Those groups are depicted in Figure \ref{fig:coal-model}.
\end{example}

\subsection{\textit{PCA} based coalition}

The main principle of a Principal Component Analysis (PCA) is to reduce a dataset to its simplest expression in terms of attributes. In other words, if the dataset is considered a multidimensional matrix, the PCA aims to reduce its dimensionality as much as possible. To do that, the different attributes of the dataset are combined linearly, the result being a new set of attributes, each new attribute being a linear combination of the previous ones.\par
Our reasoning, for this approach, is to consider the set of combined attributes (summarized by the new attribute of the PCA) as a group of influence.

Given a dataset $D = (A,X)$ composed of a set of $n$ attributes $A = \{a_1,...,a_n$\}, and a set of instances $X$ where $x \in X, x=\{x_1,...,x_n\} \forall i \in [1..n], x_i \in a_i$.

We can apply a principal component analysis which produces a new dataset $D' = (A',X')$ such as $A' = \{a'_1,...,a'_m$\} with each new attribute being a linear composition of the previous attributes : $\forall i, a'_i \in A', \exists ~  \{\alpha_1,...,\alpha_n\} \in R^n, a'_i = \alpha_1*a_1 + ... + \alpha_n + a_n$.

Each new instance is associated with an instance of the previous dataset. $\forall x' = \{x'_1,...,x'_m\} \in X', \exists! ~ x \in X, \forall i \in [1,..,m] \exists ~  \alpha_1,...,\alpha_n \in R^n, x'_i = \alpha_1*x_1 + ... + \alpha_n + x_n $.

Given this set of factors $\alpha_1,...,\alpha_n$, for each attribute, we consider each factor as an evaluation of the importance of the attributes in the group. We can then constitute a coalition of attributes by exploiting the groups formed by the most important factors. This gives us the algorithm \ref{alg:pca}. For the sake of simplicity, we consider each $a' \in A'$ as a vector of its $\alpha_i$ factors.

\begin{algorithm}[h]
\begin{algorithmic}
\Require a threshold $t$ and the set of attributes $A'$ of the PCA 
\Ensure $\sigma$ a coalition of attributes
    \State $\sigma \leftarrow \{\}$  %\Comment{initialisation of the coalition}
    \ForAll{$a' \in A'$}    \Comment{for each attribute generated by the PCA}
        \State $g \leftarrow \{\}$   \Comment{$g$, a new possible group}
        \State $\alpha max \leftarrow max(a' = \alpha_1,...,\alpha_n)$    \Comment{find the most important factor}
        \ForAll{$\alpha_i \in a'$}  
            \If{$\alpha_i \geq \alpha~max*(1-t)$}
                \State add $a_i$ to $g$ \Comment{the attribute is included in the group if close to the max}
            \EndIf
        \EndFor
        \State add $g$ to $\sigma$
    \EndFor
    \State \Return $\sigma$

\end{algorithmic}
\caption{PCA-based coalition extraction.}
\label{alg:pca}
\end{algorithm}

\begin{figure}[h]
    \centering
    \includegraphics[width = \linewidth]{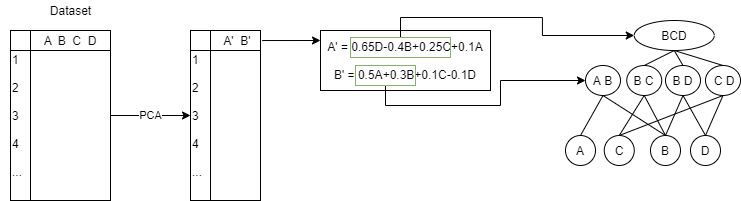}
    \caption{Depiction of the groups calculated by the PCA based coalition method for a dataset of 4 attributes. The new attributes formed by the PCA are a combination of the previous attributes. The attributes with the highest coefficient for each new attribute are considered as part of a group to be calculated.}
    \label{fig:2_coal-PCA}
\end{figure}

\begin{example}
Given our previous dataset of 4 attributes, we run a PCA on it. The new attributes generated are as in Figure \ref{fig:2_coal-PCA}. We have two principal components : $A' = 0.65D-0.4B+0.25C+0.1A$ and $B' = 0.5A+0.3B+0.1C-0.1D$. Here, we consider each attribute with the highest associated coefficients as part of a group. So, here, we have two groups: one for $A'$, $\{D,B,C\}$ and one for $B'$, $\{A,B\}$. We then calculate the total influence of each attribute as the combined influences of each attribute alone and each subgroup of the two generated groups containing the attribute. Thus, the total influence of $A$ is composed of the influence of $\{A\}$ and $\{A,B\}$ as no other group of subgroup generated contains $A$. For the total influence of $B$, we use the influence of $\{B\},\{A,B\},\{B,C\},\{B,D\}$ and $\{B,C,D\}$. The total influence of $C$ depends on the influences of $\{C\},\{B,C\},\{C,D\}$ and $\{B,C,D\}$. Finally, the influence of $D$ is constituted by the influences of $\{D\},\{B,D\},\{C,D\}$ and $\{B,C,D\}$.
\end{example}

\subsection{\textit{VIF} and \textit{revVIF} based coalition}

The variance inflation factor (VIF) is an estimation of the multicollinearity of the attributes of the dataset regarding a given target attribute.

Given a dataset $D = (A,X)$, the VIF value of $a \in A$  is calculated by running a standard linear regression with $a$ as the target for the prediction. Then, given $R$ the coefficient of determination of the linear regression, we have:
\begin{equation}
    VIF(a) = \dfrac{1}{1-R^2}
\end{equation}

It is commonly accepted that a variance inflation factor superior to 10 indicates strong multicollinearity of the attribute with other attributes of the dataset. This threshold of 10 is arbitrary but considered as a standard in numerous publications (e.g. \cite{makki2019efficient}).
Moreover, when an attribute is removed from the dataset, the VIF of the attributes multicollinear with it decreases. Then, we can automatically detect groups of attributes by calculating the VIF of each attribute (considered as a target) of the dataset, and then comparing them with a new VIF calculation with an attribute removed. For this purpose, we consider two possible approaches:
\begin{itemize}
    \item Considering as a priority the calculation of strongly multicollinear groups of attributes: Those are groups of attributes with a dependency on one another. In the context of this approach, attributes whose VIF varies strongly when an attribute is removed from the dataset is considered as part of the group.
    \item Considering as a priority the calculation of weakly or non-multicollinear groups of attributes: Given the fact that correlated attributes tend to bring the same information to the model, it may be preferable to prioritize groups for which the addition or removal of an attribute changes greatly the information brought by the group.
\end{itemize}

\begin{algorithm}[htbp]
\begin{algorithmic}
\Require a threshold $t$, the set of attributes of the dataset $A$ and a function $VIF(A)$ calculating the array of all the VIF of all the subsets of a set of attributes 
\Ensure $\sigma$ a coalition of attributes
    \State $\sigma \leftarrow \{\}$  %\Comment{initialisation of the coalition}
    \State $oldvifs \leftarrow VIF(A)$ \Comment{calculating the initial VIFs of the attributes}
    \ForAll{$a \in A$}    %\Comment{For each attribute in the dataset}
        \State $g \leftarrow \{\}$   %\Comment{initialisation of the new group of attributes}
        \State add $a$ to $g$
        \State $newvifs \leftarrow VIF(A/a)$
        \ForAll{$a' \in A$} 
            \If{$newvifs(a') < oldvifs(a')*(0.4+t)$}
                \State add $a'$ to $g$
            \EndIf
        \EndFor
        \State add $g$ to $\sigma$
    \EndFor
    \State \Return $\sigma$
\end{algorithmic}
\caption{VIF-based coalition extraction.}
\label{alg:vif}
\end{algorithm}

These two approaches are named \textit{VIF coalition} and \textit{reverse VIF coalition}, respectively. This gives us the algorithm \ref{alg:vif}, for the \textit{VIF coalition}. The \textit{reverse VIF coalition} can be obtained simply by replacing the condition for adding an attribute to a group by $if~newvifs(a') > oldvifs(a')*(1-t*0.05)$. This supplementary ratio of $0.05$ has been obtained by preliminary experiments, which showed that just keeping the $1-t$ factor led to a generation of all the possible subgroups, which defeat the principle of an approximation.

\begin{figure}[h]
    \centering
    \includegraphics[width = \linewidth]{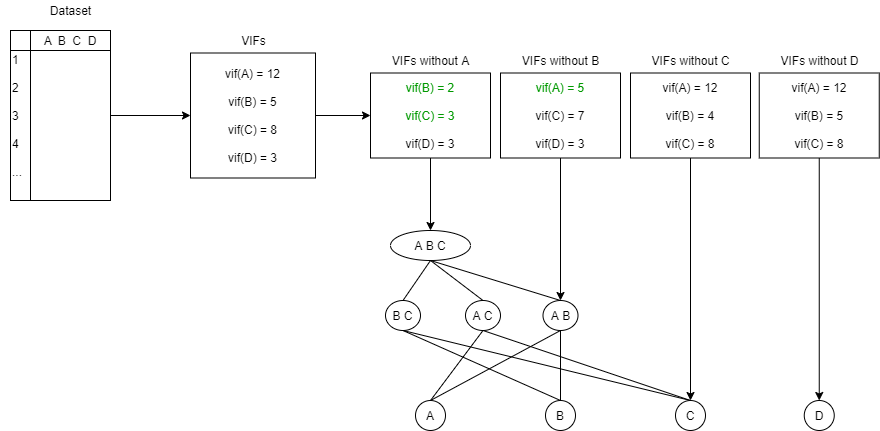}
    \caption{Depiction of the groups calculated by the VIF based coalition method for a 4 attributes dataset. the VIFs are calculated for each attribute and are recalculated with an attribute absent from the dataset. The attributes whose VIFs varies the most are considered as grouped with the removed attribute. If no VIF is changed, the removed attribute is considered as a singleton.}
    \label{fig:coal-vif}
\end{figure}

\begin{example}
Given our dataset of 4 attributes, we calculate the VIFs of each attribute. Then, we calculate the VIFs again each time with one of the attributes removed. The results are depicted in Figure \ref{fig:coal-vif}. In the case where $A$ is removed, we see that the VIFs of $B$ and $C$ vary greatly. Thus, our first group is $\{A,B,C\}$. Then, when $B$ is removed, only the VIF of $A$ varies a lot. We then have a second group: $\{A,B\}$. Finally, we can see that removing $C$ and $D$ do not make the other VIFs vary in a significant way. Because of that, the attributes $C$ and $D$ are considered as singletons. As the group $\{A,B\}$ is contained by $\{A,B,C\}$, our final coalition is $\{\{A,B,C\}\{D\}\}$. The complete influence of $A$ is constituted of the influences of $\{A\}$, $\{A,B\}$, $\{A,C\}$ and $\{A,B,C\}$. the complete influence of $B$ includes $\{B\}$, $\{A,B\}$, $\{B,C\}$, $\{A,B,C\}$. The complete influence of $C$ contains $\{C\}$, $\{A,C\}$, $\{B,C\}$, $\{A,B,C\}$. Finally, the complete influence of $D$ only contains $\{D\}$.
\end{example}

\subsection{Spearman correlation based coalition}

A limit of the variance inflation factor is the sole consideration of multicollinearity, while a correlation between attributes might not be linear. This problem is addressed through the Spearman correlation coefficient, which takes into account non-linear correlations. Spearman being not multicollinear, the calculation of the correlation between attributes has to be done by pairs. Thus, the method consists of generating the matrix of all the correlations of each pair and then deciding which attributes are part of a group. For this method, we have the same two possibilities as for the \textit{VIF} method: we can either prioritize the calculation of strongly correlated attributes or on the contrary, prioritize groups of non-correlated attributes. These two approaches are named respectively \textit{Spearman coalition} and \textit{reverse Spearman coalition}.

Given a dataset $D = (A,X)$, with $A = \{a_1,...,a_n\}$ the correlation matrix $C$ is obtained by computing the Spearman correlation coefficient of each attribute couple : $C(1,2) = corr(a_1,a_2)$. Thus $C$ is symmetrical and have 1 as the value of its whole diagonal.
For each line $i$ of the matrix $C$, we consider as grouped with $a_i$ the attributes strongly (or weakly) correlated with $a_i$, for the \textit{Spearman coalition} (or the \textit{reverse Spearman coalition}).

\begin{algorithm}[htbp]
\begin{algorithmic}
\Require a threshold $t$, the set of attributes of the dataset $A$, and a function $spearman(A)$ calculating the matrix of all the absolute Spearman correlation coefficient of all the subsets of a set of attributes. a $max$ and $min$ functions which returns the maximum and minimum of a matrix line.
\Ensure $\sigma$ a coalition of attributes
    \State $\sigma \leftarrow \{\}$  %\Comment{initialisation of the coalition}
    \State $corrmat \leftarrow spearman(A)$ \Comment{calculating the correlation matrix}
    \ForAll{$a \in A$}    %\Comment{For each attribute in the dataset}
        \State $g \leftarrow \{\}$   %\Comment{initialisation of the new group of attributes}
        \ForAll{$a \in A$} 
            \If{$corrmat(a,a')>max(corrmat(a))*(1-t)$ and $max(corrmat(a))>0.1$}
                \State \Comment{If the most correlated attribute have a coefficient less than $0.1$, we consider $a$ as a singleton}
                \State add $a'$ to $g$
            \EndIf
        \EndFor
        \State add $g$ to $\sigma$
    \EndFor
    \State \Return $\sigma$
\end{algorithmic}
\caption{Spearman-based coalition extraction.}
\label{alg:spearman}
\end{algorithm}

The algorithm \ref{alg:spearman} details the \textit{Spearman coalition} method. The \textit{reverse Spearman coalition} method can be obtained by replacing the condition for adding an attribute to a group by $corrmat(a,a')<min(corrmat(a))+max(corrmat(a))*t$ and $min(corrmat(a)) < 0.5$. This adds the least correlated attributes up to a threshold : if the attribute least correlated to $a$ have its Spearman correlation to $a$ superior to $0.5$, we consider the attribute $a$ as a singleton.

\begin{figure}[h]
    \centering
    \includegraphics[width = \linewidth]{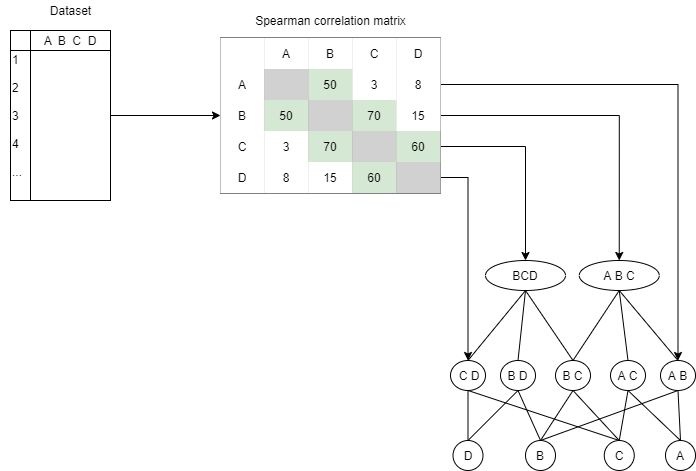}
    \caption{Depiction of the groups calculated by the spearman based coalition method for a 4 attributes dataset. The spearman correlation matrix is calculated. For each line, the attributes most correlated with the line's attribute are considered as part of a group.}
    \label{fig:coal-Spearman}
\end{figure}

\begin{example}
Given our previous dataset of 4 attributes, we calculate the matrix of the spearman correlation coefficients as depicted in Figure \ref{fig:coal-Spearman}. In this matrix, we iterate on each row of the matrix, in order to create groups based on the most correlated attributes. In the first line, we see that the attribute most correlated to $A$ is $B$, and the two other attributes are very weakly correlated to $A$. Thus, we have a first group: $\{A, B\}$. The second line tells us that $A$ and $C$ are both strongly correlated to $B$. So, we have a second group: $\{A,B,C\}$. Similarly, the third line indicates that $B$ and $D$ are correlated to $C$, and so we add a third group: $\{B,C,D\}$. Finally, by looking at the last line we learn that only $C$ is strongly correlated to $D$ and so our last group is $\{C, D\}$. As the two groups of cardinal 2 are contained by the two groups of cardinal 3, we have our final coalitions: $\{$\{A,B,C\}$,$\{B,C,D\}$\}$. With this coalition, the complete influence of the attribute $A$ is composed of $\{A\}$, $\{A,B\}$, $\{A,C\}$ and $\{A,B,C\}$. $B$ is composed of $\{B\}$, $\{B,D\}$, $\{A,B\}$, $\{B,C\}$, $\{A,B,C\}$ and $\{B,C,D\}$. Finally, the complete influence of $D$ is composed of $\{D\}$, $\{C,D\}$, $\{B,D\}$ and $\{B,C,D\}$. 
\end{example}

%\subsection{Choosing a specific alpha threshold for each coalition method}

\section{Evaluation of the additive methods} 
\label{sec:experiments}
%PENSER A AJOTER LES TESTS SVM/RANDOM FOREST

In this section we aim to evaluate the performances of each coalition calculation method, considering their precision when compared to the \textit{complete} influence, and their computational time.
We also give an overview of the group characterization for each coalition method.

\subsection{Experimental protocol}

Our experiments are run on an AMD Ryzen 3700 processor with 8 x 3.6 GHz cores and 32 GB of RAM.
Our tests are realized from the data available on the Openml platform \cite{Vanschoren2013}.  We select the biggest collection of datasets \footnote{Available in \url{https://www.openml.org/s/107/tasks}} on which classification tasks have been run. We also consider two classification tasks: Random Forest and Support Vector Machine (SVM) with the non-linear Radial-Basis-Function (RBF) kernel. Experiments are conducted using Python 3.7.9 with the Scikit-Learn package on both models \footnote{\url{https://scikit-learn.org/stable/}}. Due to the heavy computational cost of the complete influence -considered as the reference of our experiments- we select the datasets having at most nine attributes.
Thus, a collection of 243 datasets is obtained. Considering the two types of workflows, we have a total of 486 runs.
For each of those runs, we generate each type of influence described in this paper, for each instance of the 243 datasets: the \textit{complete} influence for the baseline, along with the \textit{coalitional} influences, \textit{k-depth} influence, and the Kernel and Tree SHAP influences. 
The \textit{coalitional} influences are generated using the different group generation methods described in Section \ref{sec:new_methods}, which are based on an $\alpha \in ]0,0.5[$ parameter (small values of  $\alpha$ resulting in smaller subgroups, and high values in bigger ones). We generate the possible subgroups with 5 different values of  $\alpha$ to study the influence of subgroup size.
To compare the different explanation methods, we consider the explanation results as a vector of attribute influences noted $\mathcal{I}(x) = [i_1,...,i_n]$ with $n$ the number of attributes in the dataset. Thus, each of the attributes $a_k$ is given an influence $i_k \in [0,1]$ by the method $\mathcal{I}$ : $\forall k \in [1..n], i_k = \mathcal{I}_{a_{i}}(x)$, with $x$ an instance of the dataset.
We then define a difference between two vectors of influences $i,j$ as the normalised Euclidean distance: 
\begin{equation}
d(i,j) = \frac{1}{2\sqrt{n}} \sum_{k=1}^{n} \sqrt{(i_k - j_k)^2}
\label{eq:influence_diff}
\end{equation}
Considering this formula, we define an error score based on the difference between an explanation method and the \textit{complete} influence method. Given an instance $x$, an explanation method $\mathcal{I}(x)$, and the \textit{complete influence} method $\mathcal{I}^{C}(x)$:
\begin{equation}
err(\mathcal{I},x) = d(\mathcal{I}(x),\mathcal{I}^C(x))
\label{eq:influence_diff_error}
\end{equation}
For each instance of each dataset, we generate the error score of every method, allowing us to compare their performances across the different collected datasets. Each error score is the distance of one of the coalitional methods from the \textit{complete} method. Thus, lesser error is indicative of a more precise estimation of the \textit{complete} method.\\

To compare methods, we also consider the time needed to explain a set of data, called computation time. This includes the time to determine the subgroups of interest and to compute the influences of all instances in the \textit{coalitional} and \textit{k-depth} methods. For the \textit{SHAP} methods, the calculation time is the one taken to calculate the influences of all instances.
Table \ref{tab:stats_datasets} details the number of datasets and mean number of instances for each number of attributes. Since the number of instances impacts the total computation time for a dataset, each computational time is normalized by dividing by the number of instances in the dataset to compare times per instance.

\begin{table}
    \centering
\begin{tabular}{|l|r|r|r|r|r|r|r|r|r|}
\hline
Number of attributes &    1 &    2 &     3 &    4 &    5 &    6 &    7 &    8 &    9 \\
\hline
Number of datasets       &    3 &   21 &    44 &   25 &   38 &   26 &   34 &   28 &   24 \\
Mean number of instances &  724 &  736 &  1688 &  560 &  843 &  600 &  456 &  750 &  479 \\
\hline
\end{tabular}
    \caption{Dataset stats with a given number of attributes}
    \label{tab:stats_datasets}
\end{table}

\subsection{Evaluation of the literature methods}
\label{sec:time-error-scores-litterature}

\begin{figure*}
    \centering
    \includegraphics[width=0.75\textwidth]{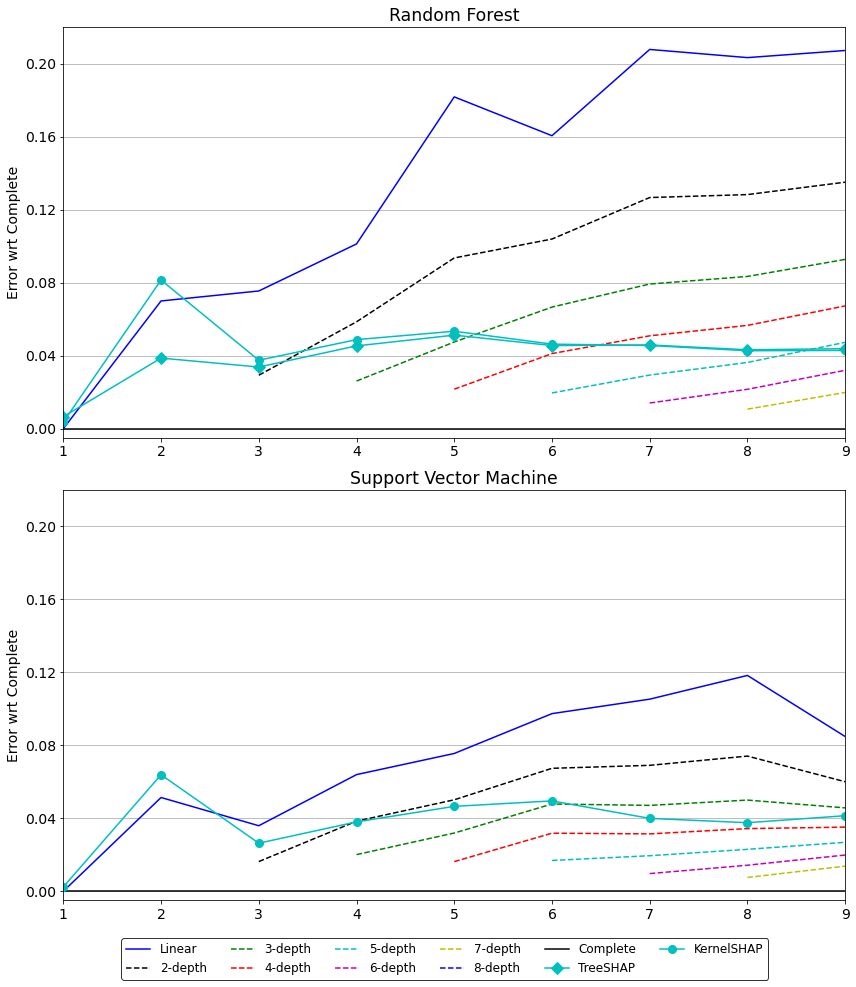}
    \caption{Error score between each explanation method and the \textit{complete influence} depending on the number of attributes in the dataset.}
    \label{fig:litterature_performance_error}
\end{figure*}

In this section, we focus on the \textit{k-depth} and \textit{SHAP} methods and evaluate them with the protocol described previously. Figure \ref{fig:litterature_performance_error} indicates the mean error with respect to the \textit{complete} method, and with the distinction of the datasets based on their number of attributes. For both models, the \textit{linear} (or 1-depth) method gives the worst results, particularly as the number of attributes grows. A larger \textit{k} in \textit{k-depth} methods results in more precise influence attribution.
This is fully expected since a higher $k$ generates larger groups closer to the $Complete$ method. Interestingly, \textit{SHAP} methods -\textit{KernelSHAP} for both models and \textit{TreeSHAP} only for the Random Forest- give rather accurate results, without really losing accuracy as the number of attributes grows.\\

\begin{figure*}
    \centering
    \includegraphics[width=0.75\textwidth]{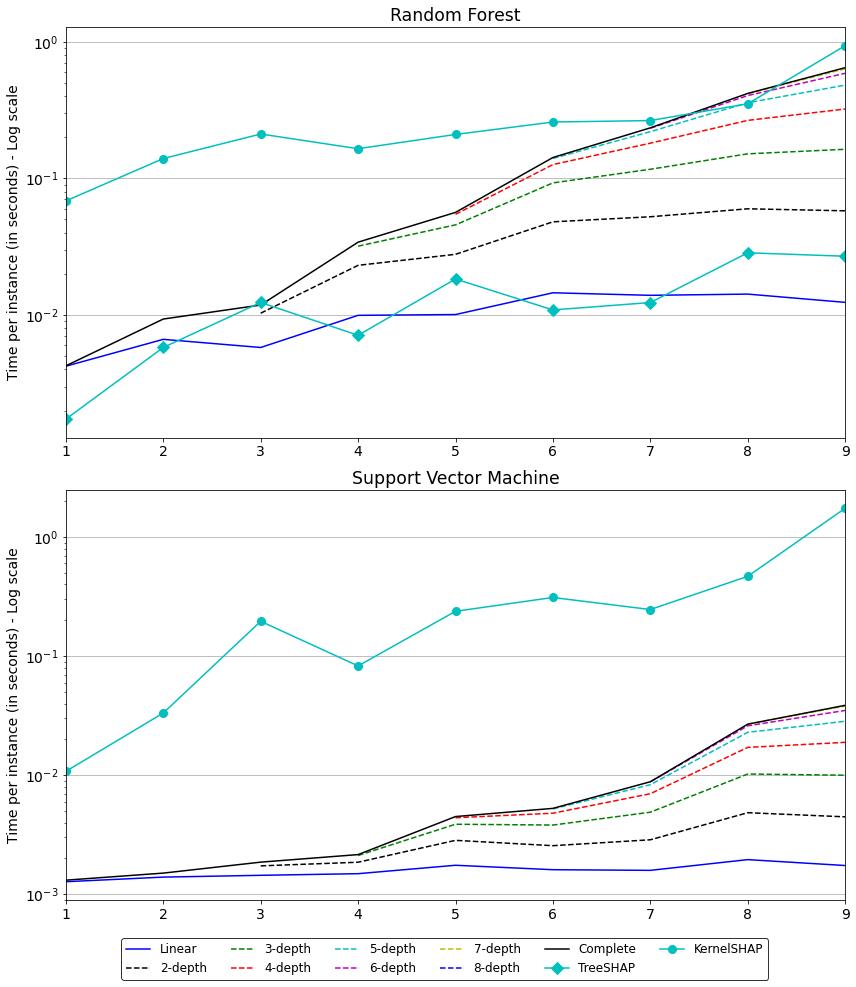}
    \caption{Time per instance (in seconds and log scaled) taken by each method depending on the number of attributes in the dataset.}
    \label{fig:litterature_performance_time}
\end{figure*}

Figure \ref{fig:litterature_performance_time} represents the mean computation time per instance for all methods. The y-axis is log scaled in order to have a clear view of all results, even for datasets with a low number of attributes, as the growth is exponential. The \textit{k-depth} methods take more and more time as the \textit{k} grows -from \textit{linear} to \textit{complete}. The \textit{TreeSHAP} method, only usable with the Random Forest, gives rather good computation time, being between those of \textit{linear} and \textit{2-depth} methods.
On the contrary, the \textit{KernelSHAP} methods give poor results, being slower than the complete, especially with the SVM model. This is a major inconvenience since it suggests that interpretability with models that are not tree-based would often be intractable in practice.

We then present, in the next section, the evaluations of our coalitional methods, including the first comparison with literature approaches.

\subsection{Evaluation of the coalitional methods}
\label{sec:time-error-scores-coalitional}

\begin{figure}[h]
    \centering
    \includegraphics[width=1.0\textwidth]{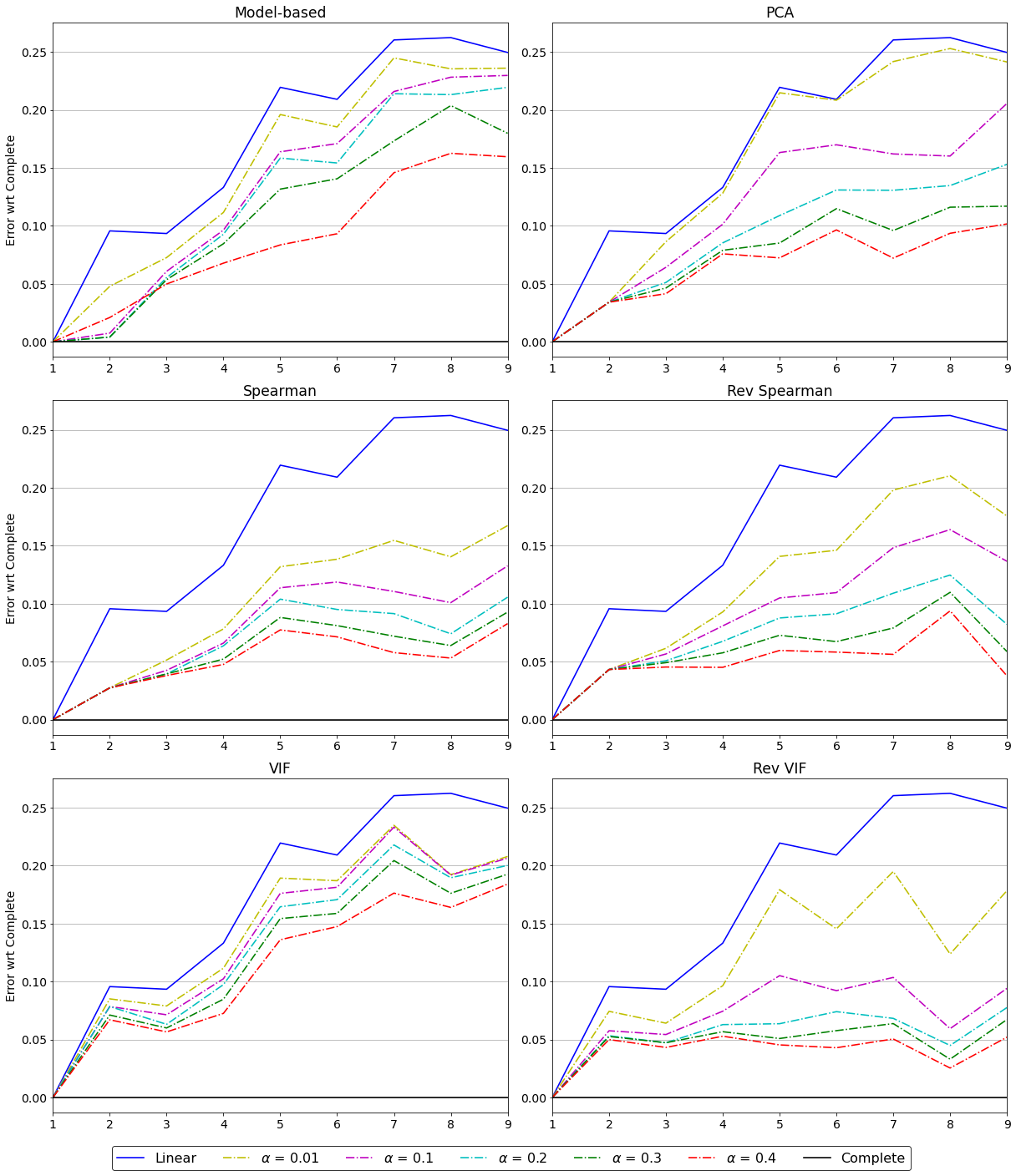}
    \caption{Error score between each coalitional method and the complete influence, versus the number of attributes in the dataset averaging over the two models}
    \label{fig:coalitional_performance_error}
\end{figure}

The error rate, with respect to the $complete$ method, of the coalitional methods (for several $\alpha$-thresholds), along with the \textit{linear} method, is shown in Figure \ref{fig:coalitional_performance_error}. All methods give better results with a higher $\alpha$-threshold since a larger one generates bigger subgroups, thus leading to higher complexity.
In particular, $RevVIF$ approach seems to give better results compared to the other ones, for a majority of the  $\alpha$-thresholds. A possible reason is the ability of $RevVIF$ to generates groups of "less correlated" attributes. We can suppose these types of groups better represent the diversity of possible explanations.
\\

\begin{figure}[h]
    \centering
    \includegraphics[width=1.0\textwidth]{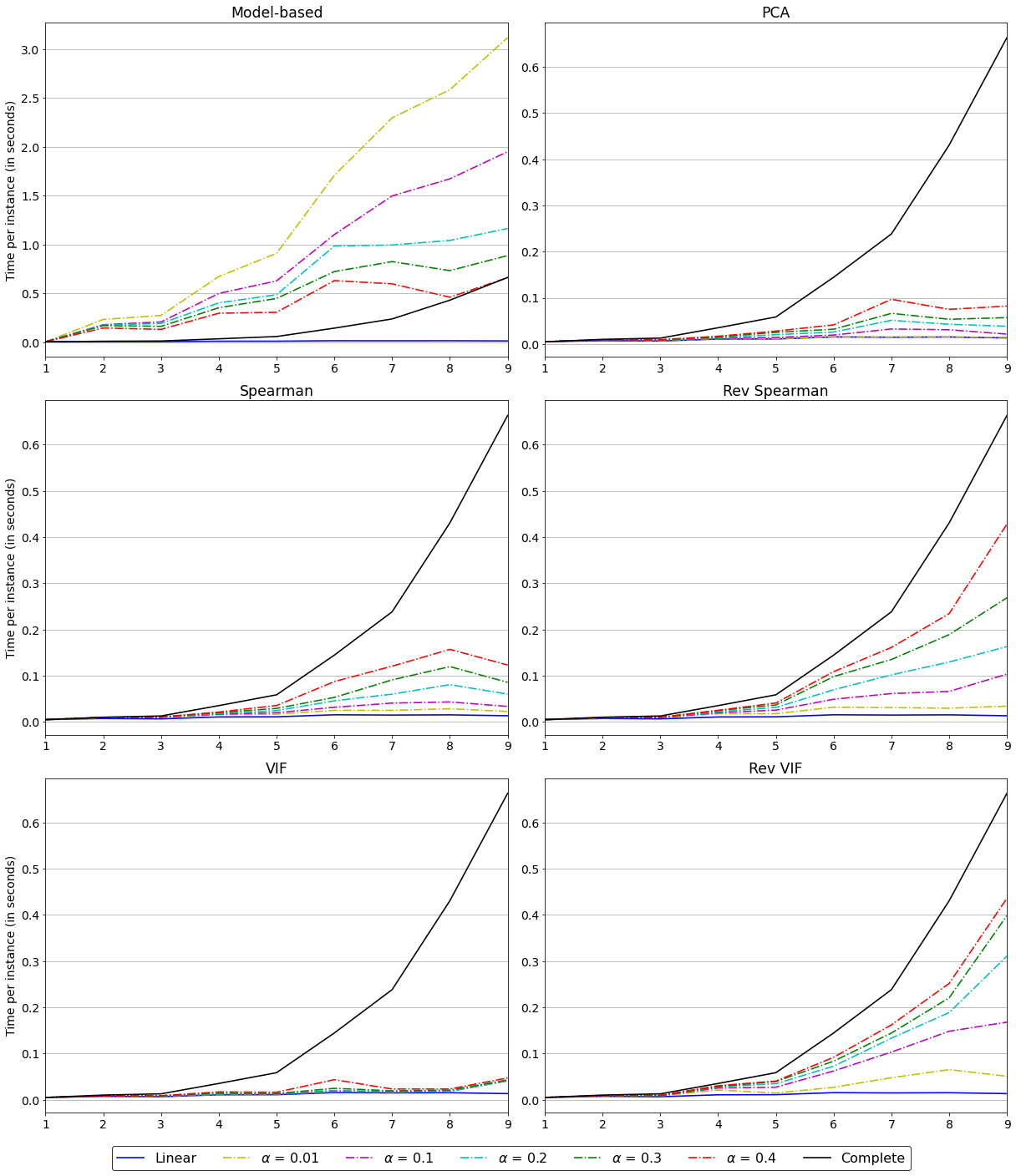}
    \caption{Time per instance of each coalitional method versus the number of attributes in the dataset, averaging over the two models}
    \label{fig:coalitional_performance_time}
\end{figure}

Figure \ref{fig:coalitional_performance_time} shows computation time per instance for each coalitional grouping method and the complete one. All methods except for \textit{Model-based} are notably faster than the \textit{complete} method, especially for \textit{PCA}, \textit{Spearman} or \textit{VIF}. On the contrary, \textit{Model-based} takes mostly more time than the \textit{complete} method. 
Indeed, the time taken by the \textit{Model-based} method to generate a set of groups is particularly long compared to other coalition-based methods, thus canceling out the effects of group selection. This automatically disqualifies this method for practical use.
In particular, the computation time for $RevVIF$ generally seems higher than the other methods. A possible reason is that it should generate larger groups, closer to the $Complete$ one.

\begin{figure*}
    \centering
    \includegraphics[width=0.75\textwidth]{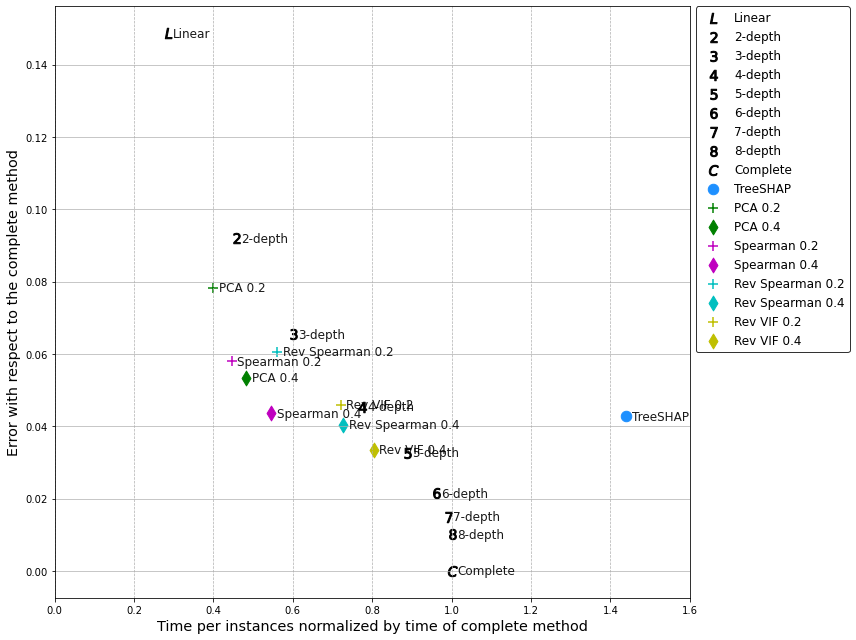}
    \caption{Performance map over all datasets for k-complete, SHAP and coalitional methods for Random Forest}
    \label{fig:coalitional_performance_map_Random_Forest}
\end{figure*}

\begin{figure*}
    \centering
    \includegraphics[width=1\textwidth]{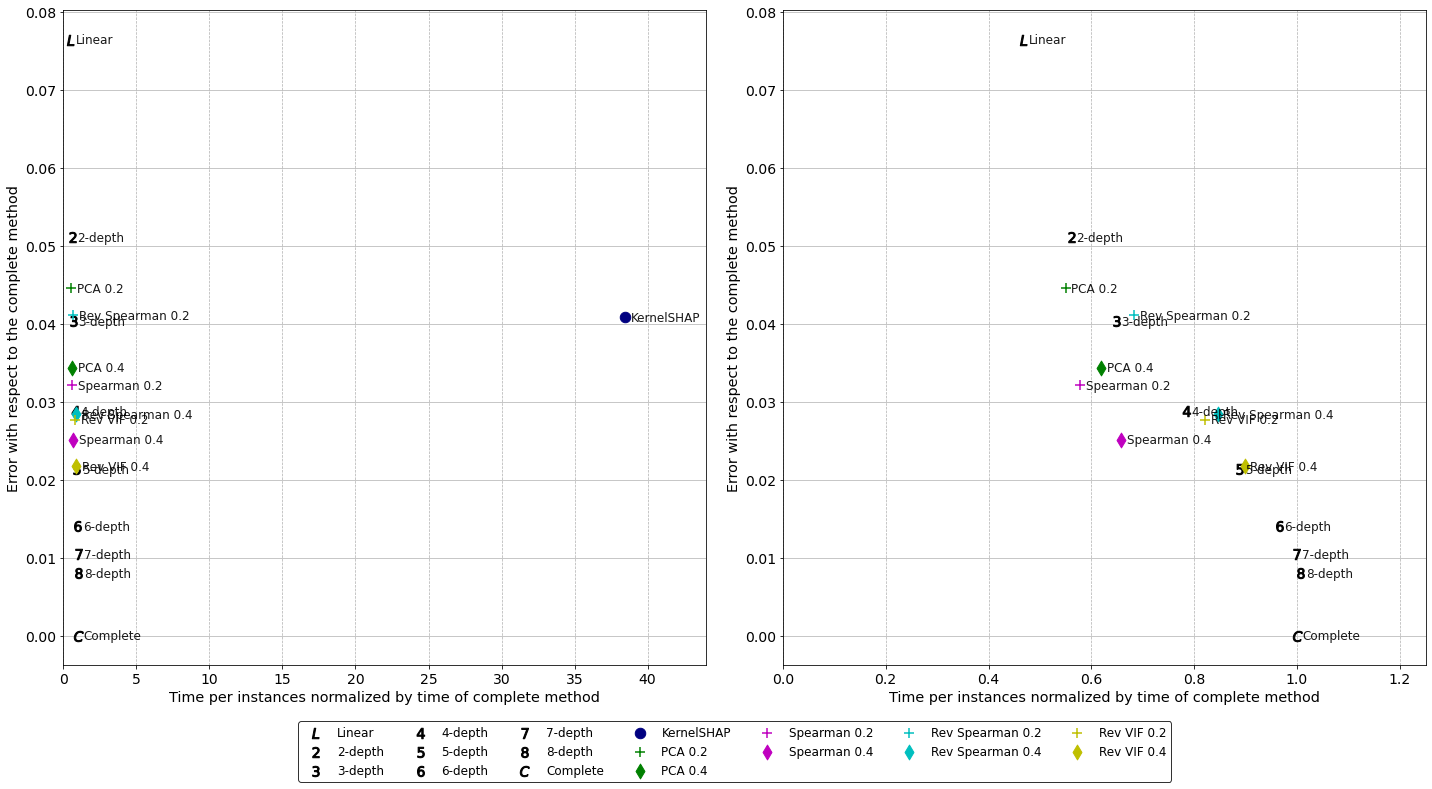}
    \caption{Performance map over all datasets for k-complete, SHAP and coalitional methods for SVM}
    \label{fig:coalitional_performance_map_SVM}
\end{figure*}

To have a clearer view of the methods' performances, we average the error with respect to the $Complete$ method and the computation time globally, thus independently of the number of attributes in the datasets. Therefore, it gives us a single representation, called \textit{Performance Map} (Figures \ref{fig:coalitional_performance_map_Random_Forest} and \ref{fig:coalitional_performance_map_SVM}), with the computation time normalized by one of the complete on the horizontal axis, and the error for the complete on the vertical axis. The complete method is thus placed on the point with coordinates $(1, 0)$. All the methods are thus placed above the complete since they can not have a null error with respect to the complete, and methods placed at the left of the complete have lower computation time than the complete, while those placed at the complete right are slower to compute.

We also retained only the most promising grouping methods from previous results -\textit{PCA}, \textit{Spearman}, \textit{Reverse Spearman} and \textit{Reverse VIF}- for $\alpha \in [0.2, 0.4]$ so as to avoid a figure with too much noisy information. The \textit{k-depth} and $SHAP$ methods are also shown.

Figure \ref{fig:coalitional_performance_map_Random_Forest} shows the \textit{Performance Map} for the Random Forest model. \textit{KernelSHAP} is not included since its computation time is too high, thus flattening the rest of the graph. Nevertheless, \textit{TreeSHAP} method is still on average slower to compute than the complete, despite being specially designed for tree-based models. As for the coalitional methods, \textit{PCA} is the fastest, but not the most accurate, while \textit{Reverse VIF} is the most accurate but also the slowest method between all coalitional methods. \textit{Spearman} and \textit{Reverse Spearman} seems to be the best balance between precision and computation time.

Figure \ref{fig:coalitional_performance_map_SVM} shows the \textit{Performance Map} for the SVM model. The subfigure on the left includes the \textit{KernelSHAP} method, but as this method is on average 40 times slower than the $Complete$ method for our datasets, it flattens the other ones. Thus the subfigure on the right does not include \textit{KernelSHAP} method. In a similar fashion to Random Forest, \textit{PCA} is the fastest coalitional method but the least accurate, while \textit{Reverse VIF} is the most accurate. Again, \textit{Spearman}-based grouping methods seem to be the most balanced ones.\\

The only advantage of \textit{SHAP} methods is that one does not need to retrain any model. \textit{SHAP} simulates missing attributes through a heavy number of perturbations of the to-be-explained instance, implying a substantial cost. While our methods, only retraining once the model for every coalitional group, seems to be beneficial over \textit{SHAP}. That is why the \textit{complete} method, which has the highest number of coalitional groups, is often faster to compute than \textit{SHAP} methods.

The next section gives an overview of the characteristics of the groups generated by our coalitional methods. It justifies the very interesting results obtained by the $Spearman$-based methods, in particular.

~\newline

\subsection{Group characterization of coalitional methods}
\label{sec:group_characterization}

%nombre moyen de groupes
%taille moyenne des groupes
%-- pour différents alpha
%complexité des groupes (10%, 20% de la complète)

\begin{figure}[h]
    \centering
    \includegraphics[width=0.8\textwidth]{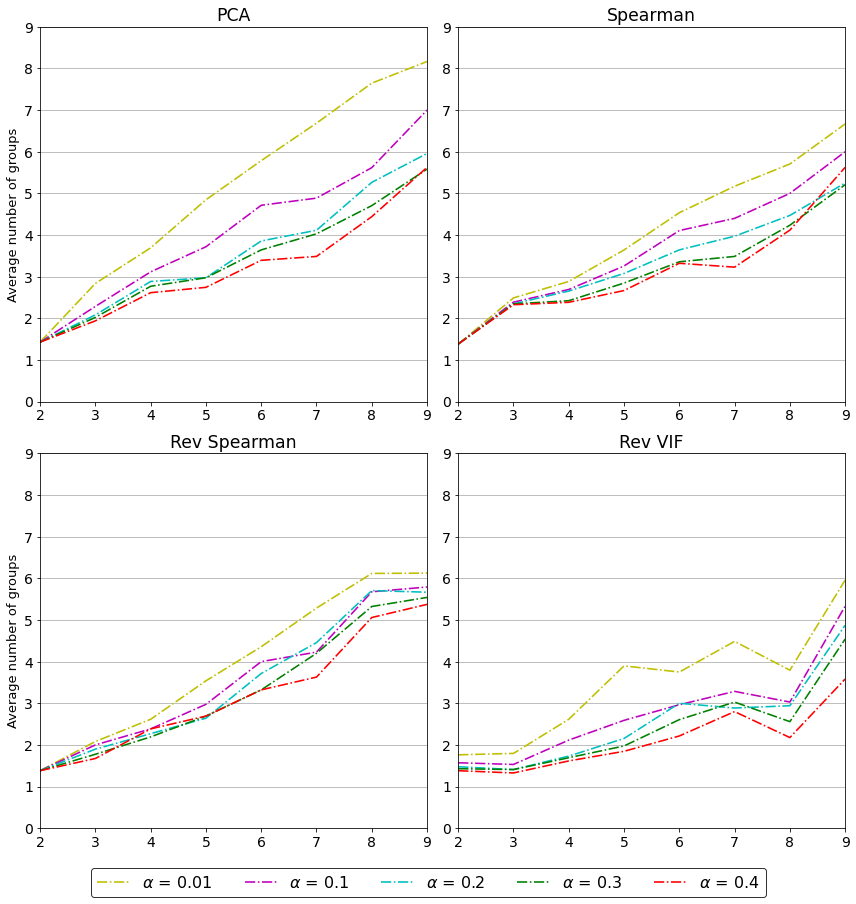}
    \caption{Mean number of groups for coalitional methods depending on $\alpha$-threshold and number of attributes}
    \label{fig:coalitional_mean_nb_groups}
\end{figure}

\begin{figure}[h]
    \centering
    \includegraphics[width=0.8\textwidth]{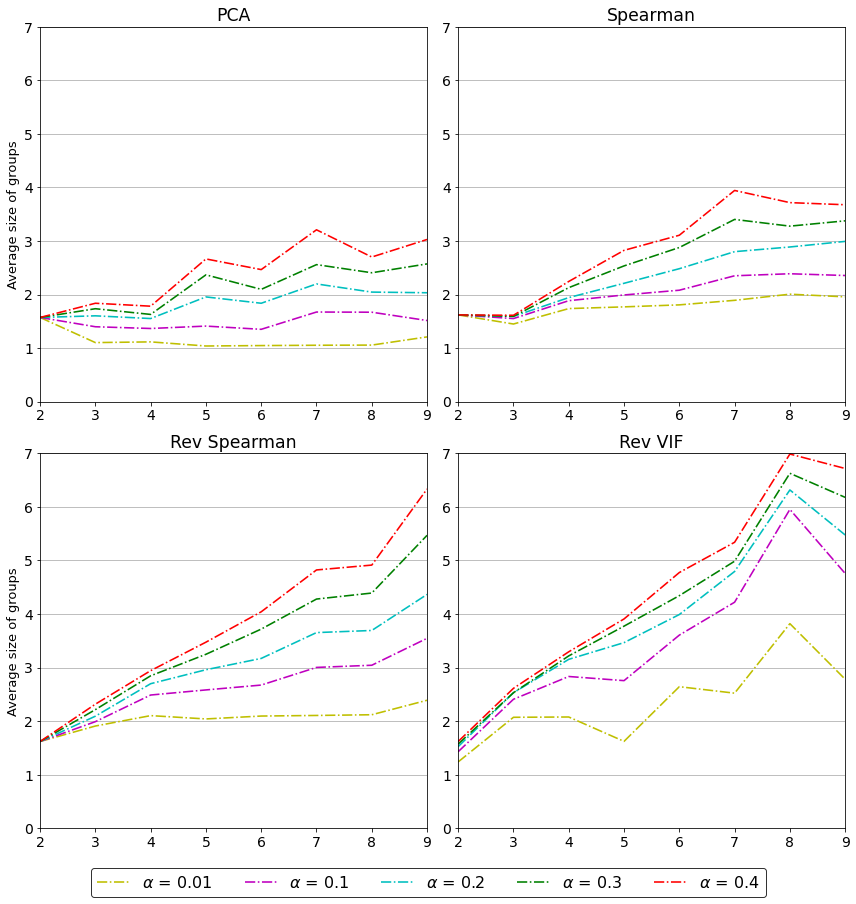}
    \caption{Mean size of groups for coalitional methods depending on $\alpha$-threshold and number of attributes}
    \label{fig:coalitional_mean_size_subgroups}
\end{figure}

%ajouter figure sur complexité alpha vis à vis de complèete et montrer, par ensemble de datasets avec peu d'instances d'une part et beaucoup d'instances d'autre part, l'impact de temps
Figure \ref{fig:coalitional_mean_nb_groups} and \ref{fig:coalitional_mean_size_subgroups} compare the average number and average size of the groups of attributes generated by coalitional methods for several $\alpha$-thresholds. We see that the grouping method and the $\alpha$-threshold parameter have an important impact on these metrics, thus on the complexity of the groups. \textit{Reverse VIF} generates the highest group sizes on average, which can explain the fact that this method gives the lowest error with respect to the complete, but also the longest computation time as discussed in Section \ref{sec:time-error-scores-coalitional}. On the contrary, \textit{PCA} gives the highest number of groups along with the lowest size of groups explaining the fact that this method is the fastest since it generates a list of groups closer to the \textit{linear} method than the \textit{complete}.\\

\begin{figure}[h]
    \centering
    \includegraphics[width=0.8\textwidth]{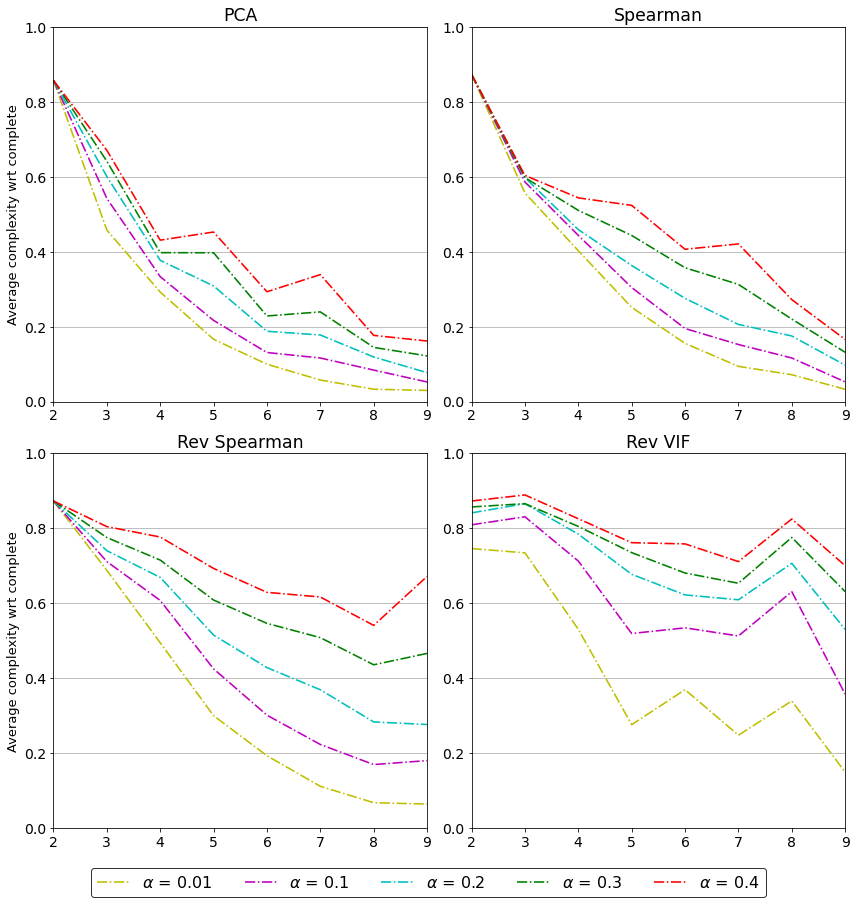}
    \caption{Mean complexity proportion with respect to complete complexity for coalitional methods depending on $\alpha$-threshold and number of attributes}
    \label{fig:coalitional_complexity_proportion}
\end{figure}

Figure \ref{fig:coalitional_complexity_proportion} shows the average complexity of the generated groups with respect to the complexity of the \textit{complete} method which is equal to $2^{nb\, attributes} - 1$. The complexity of a list of groups generated by a grouping method is equal to the number of distinct subgroups. The group complexity is then divided by the \textit{complete} one to get a value between 0 and 100\%, the latter indicating that the generated group list is similar to the \textit{complete}, while a low value indicates that the group list is closer to the \textit{linear}, thus less complex and faster to compute. As expected, the \textit{PCA} method generates on average the least complex groups, while the \textit{Reverse} methods, whether based on \textit{Spearman} or \textit{VIF}, generate the most complex ones (particularly true for $RevVIF$) and explaining a higher computation time.

\begin{figure}[h]
    \centering
    \includegraphics[width=0.65\textwidth]{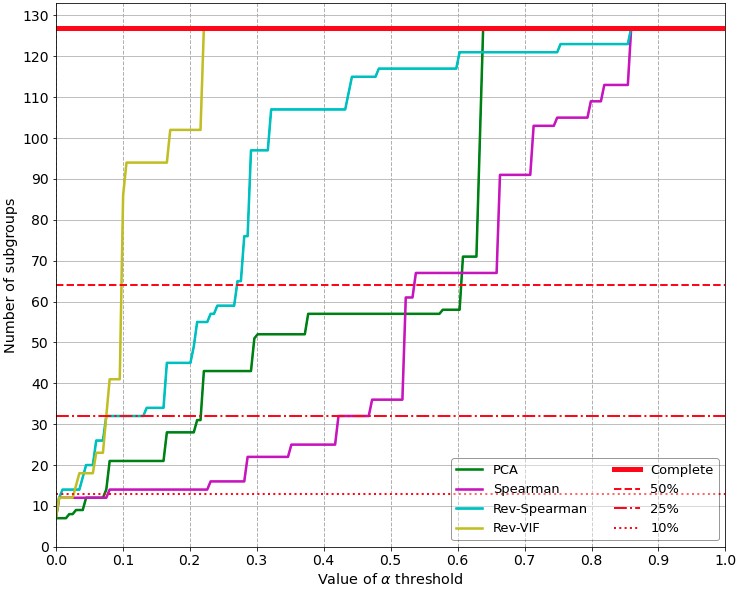}
    \caption{Evolution of complexity of coalitional methods depending on $\alpha$-threshold}
    \label{fig:coalitional_nb_subgroups_evolution_wrt_alpha}
\end{figure}

The previous results on the group characterization for coalitional methods show that the groups differ greatly in function of the $\alpha$-threshold and the grouping method. Figure \ref{fig:coalitional_nb_subgroups_evolution_wrt_alpha} displays the evolution of the complexity of the groups generated by four coalitional methods for a particular dataset with 7 attributes. The \textit{linear} complexity is thus equal to $7$ while the \textit{complete} complexity is $2^7-1=127$. There is a clear difference between the evolution of the grouping methods confirming that the $\alpha$-threshold can not be set at the same value for all methods. In this example, if one wants a complexity which equals to 25\% of the \textit{complete}, the $\alpha$-threshold for the \textit{Reverse VIF} method would be equal to about 0.08, whereas for \textit{Spearman} it would be about 0.42.

To have a more appropriate and fair comparison of the several coalitional methods, we decide not to rely only on the $\alpha$-threshold but rather on the proportion of the \textit{complete} complexity needed as shown in Figure \ref{fig:coalitional_nb_subgroups_evolution_wrt_alpha}. If one does want a short computation time they can set the proportion to 10\%, while 50\% can be set if the computation time is not a problem and that more accurate results are needed.
A well-used bisection method is applied to find the $\alpha$-threshold that matches the selected proportion of the \textit{complete} complexity.

In the following Section \ref{sec:comparison_literature_coalitional}, we propose a set of tests considering a complexity proportion to 10\%, 25\% and 50\% for each coalitional method. The time taken by the bisection method is, thus, included in the computation time.
It is also worth to note that for all the coalitional methods -except for the \textit{Model-based}- the group generation relies only on the $\alpha$-threshold and not on the model used. Thus one does not need to generate again the groups if we switch to another model, while any \textit{SHAP}-based method needs to be retrained for any new model, even if it only differs from the previous one from a small change in one hyper-parameter.

\subsection{Performance between literature and coalitional methods over different types of datasets}
\label{sec:comparison_literature_coalitional}

So as to have a clearer view of the impact of all methods in one figure, \textit{Performance Map} are displayed for each model. But as the number of attributes and instances in a dataset has a strong impact on the performances, we split the datasets into two parts. The first one includes datasets that have relatively few instances (strictly less than 500) or attributes (strictly less than 6), whereas the second part only includes datasets with a higher number of instances (at least 500) and attributes (at least 6). There are 213 datasets in the first set and 30 in the second one.

\begin{figure}[h]
    \centering
    \includegraphics[width=1\textwidth]{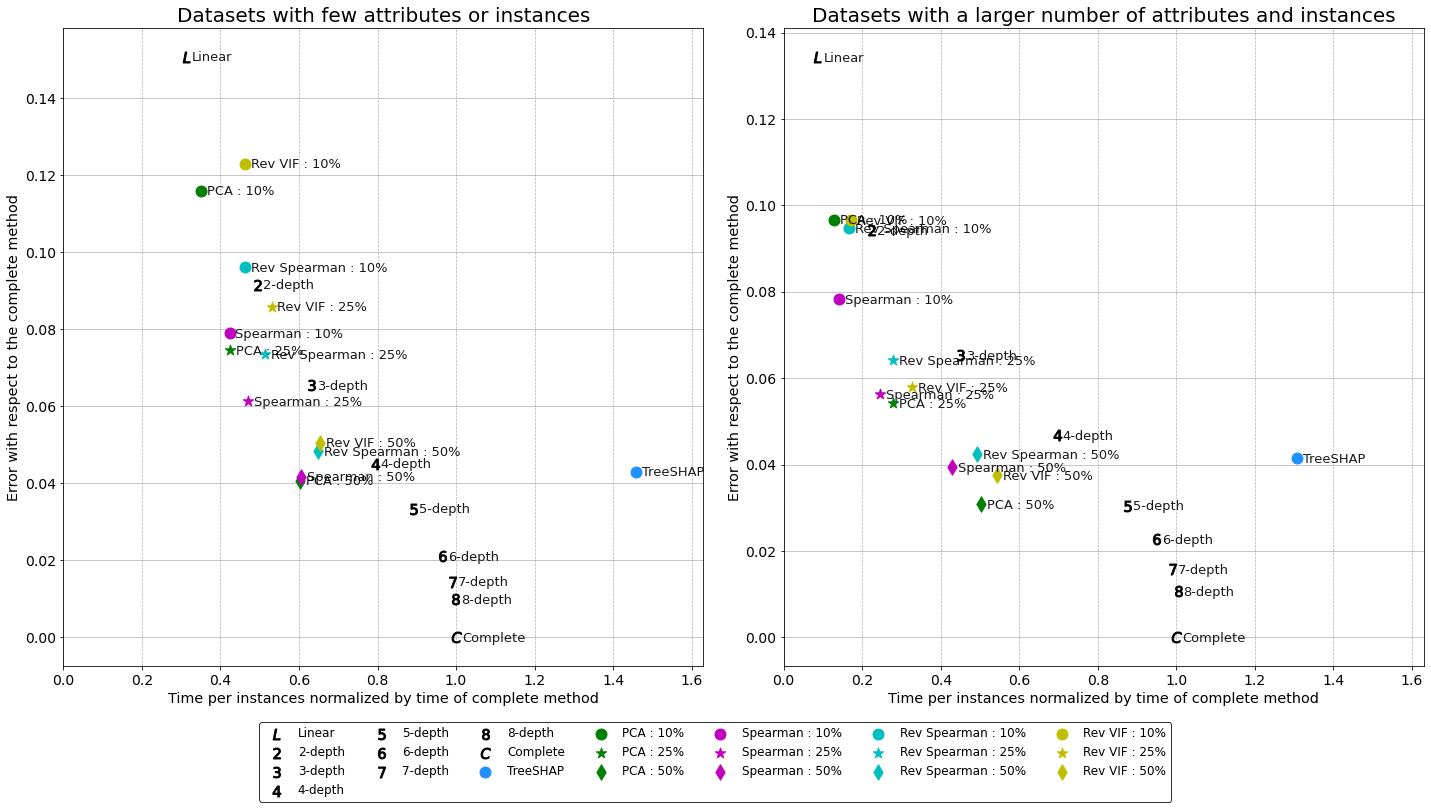}
    \caption{Performance maps for two sets of datasets for coalitional methods for Random Forest}
    \label{fig:coalitional_performance_map_final_RandomForest}
\end{figure}

Figure \ref{fig:coalitional_performance_map_final_RandomForest} shows the results for the Random Forest model for the coalitional methods -\textit{PCA}, \textit{Spearman}, \textit{Reverse Spearman} and \textit{Reverse VIF}- for three complexity proportion -10\%, 25\% and 50\%-, the \textit{k-depth} methods -from \textit{linear} to \textit{complete}- and \textit{TreeSHAP}. The left subgraph shows the results for the first set containing datasets with few attributes or instances and the subgraph on the right indicates results for more complex datasets as described previously.

\textit{TreeSHAP} has on average a higher computation time than the \textit{complete} method with an error between the \textit{4-depth} and \textit{5-depth} for both sets of datasets. Coalitional methods show strong results, \textit{PCA - 50\%} has a computation time equivalent to a \textit{3-depth} but has an error closer to the \textit{5-depth} for both sets. Similarly, all coalitional methods with a threshold of 25\% are in terms of computation time closer to the \textit{2-depth} while being more accurate than \textit{3-depth}. This is explained by the fact that coalitional methods generates smarter groups with less useless or redundant information than \textit{k-depth}, thus being more efficient.\\

\begin{figure}[h]
    \centering
    \includegraphics[width=1\textwidth]{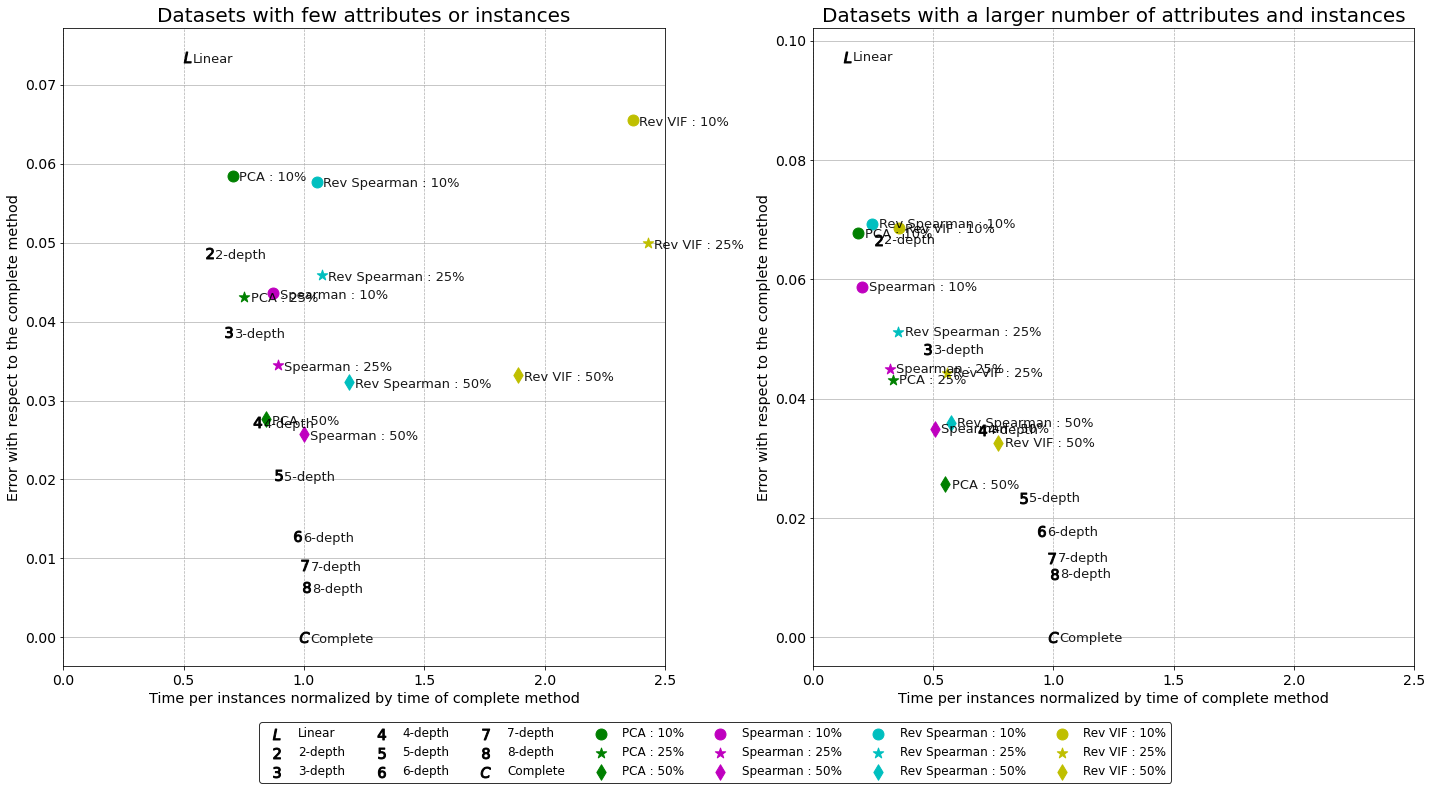}
    \caption{Performance maps for two sets of datasets for coalitional methods for SVM}
    \label{fig:coalitional_performance_map_final_SVM}
\end{figure}

Figure \ref{fig:coalitional_performance_map_final_SVM} shows the \textit{Performance Maps} for the SVM model for both sets of datasets. \textit{SHAP} are not displayed since \textit{TreeSHAP} is not usable with SVM and \textit{KernelSHAP} has a really high computational time as previously shown in Figure \ref{fig:coalitional_performance_map_SVM}, hence flattening the figure if displayed.\\
Unlike Random Forest, there is a clear difference between the results for the two sets. For smaller datasets -either in terms of an attribute or instance number- some coalitional methods are longer to compute than the \textit{complete} one. This is because the time taken to find the appropriate $\alpha$-threshold with the bisection method is too large relative to the global computation time, which also includes training models and influence computation for each attribute.

Nevertheless, for larger datasets, the performances of coalitional methods are satisfying. Indeed, in a similar way to Random Forest results, coalitional methods with a 50\%-threshold are faster to compute than \textit{4-depth} method for the same error with respect to the \textit{complete} -especially \textit{PCA} which is really efficient. \textit{Spearman} and \textit{PCA} with a 25\% threshold are very efficient as well, with a computation time between those of \textit{2-depth} and \textit{3-depth} while being more accurate than \textit{3-depth}.

\subsection{How to interpret the coalitional explanations}

In this section, we show an example of the use of our coalitional methods on a real use case dataset and for specific instances from this dataset. 
For this experiment, we only consider the \textit{Spearman} method, since it produces one of the best results as discussed in the previous sections. We also include the \textit{Kernel SHAP} method for a comparison with the explanations found with the \textit{Spearman} approach.

The quality of an interpretability method is a subjective concept and it would be difficult to theorize measures to assess what constitutes good interpretability. Nevertheless, some criteria exist in the literature to evaluate individual explanations \cite{robnik-sikonja_perturbation-based_2018}. Properties such as fidelity and comprehensibility can help non-experts to evaluate and compare individual explanations, thus explanations methods. Fidelity represents the ability of an explanation to approximate the prediction of the "black box" model and comprehensibility evaluates the ability of users to understand the explanations.

The use case dataset concerns the SARS-COV2 - also called Covid-19 - epidemic outbreak in France during spring 2020. Data collection complied with the European GDPR rules and consists of anonymized medical information of 409 patients with Covid-19 virus hospitalized at the Centre Hospitalier Intercommunal de Créteil \footnote{\url{https://www.chicreteil.fr/}} between March and May 2020.
The primary binary outcome consists of the deterioration of the patient's state of health during their stay, also called aggravation. Deterioration was defined as the requirement for mechanical ventilation, presence of septic shock, acute respiratory distress syndrome, a requirement for resuscitation maneuvers during hospitalization or hospital mortality. Out of the total number of patients, 176 of them had a deterioration in their health state, i.e. 43\% of the data set.
Each patient profile is established at the patient's arrival at the hospital. Available information consists of 10 attributes such as basic characteristics (age and gender), exam results of Chest Computed-Tomography (CT) scan severity, and comorbidities like cancer, type-2 diabetes, obesity, intellectual disability, cardiovascular disease. For this use case, a \textit{Random Forest} model and the \textit{Spearman} coalitional method with a complexity threshold of 25\% are used. The model has an accuracy of 74\% with an 80\% precision and a 69\% recall.

\begin{figure}[h]
  \centering
  \includegraphics[width=\textwidth]{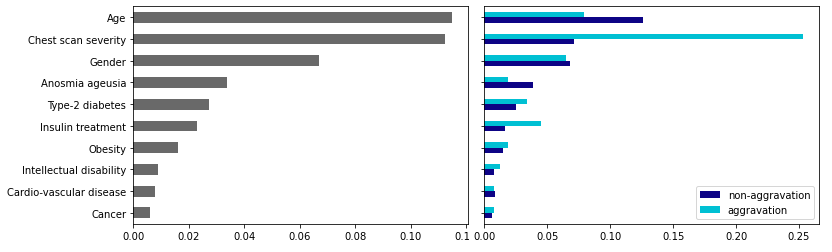}
  \caption{Mean absolute influence for each attribute with Spearman 25\% method}
  \label{fig:mean_infs_spearman25}
\end{figure}

\begin{figure}[h]
  \centering
  \includegraphics[width=\textwidth]{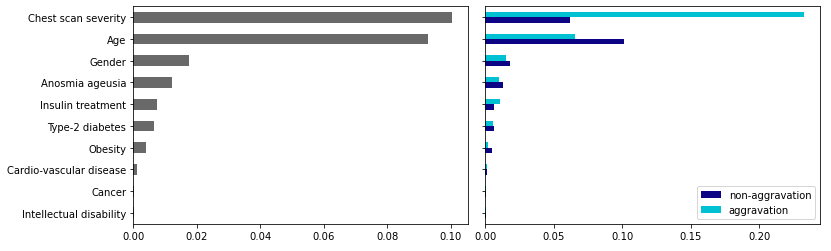}
  \caption{Mean absolute influence for each attribute with Kernel SHAP method}
  \label{fig:mean_infs_kernelshap}
\end{figure}

Figure \ref{fig:mean_infs_spearman25} and \ref{fig:mean_infs_kernelshap} give the average absolute influence of each attribute, with or without taking into account the class predicted by the model, for the \textit{Spearman 25\%} and \textit{Kernel SHAP} method respectively. Age and chest Chest scan severity are the two most important attributes for both methods, with Chest scan severity having a greater impact on aggravation class. This shows a coherence between the medical reality and both explanation methods. Indeed, a high Chest scan severity is strongly associated with an aggravation of the health state as shown in \cite{francone2020chest}. 
Both methods also have different results for other attributes, such as cardio-vascular disease, cancer and mental disability that have on average almost no impact with $Kernel SHAP$ and all attributes have on average a higher influence with the $Spearman$ method.
Taking into account classes, the average influences for both classes are relatively similar using $Kernel SHAP$, except for the age and severity of the chest CT scan. With the $Spearman$ method, the average influences of ageusia anosmia, diabetes and insulin treatment are dissimilar. For older patients with high chest scanner severity, type-2 diabetes, insulin treatment, or ageusia anosmia, the model is likely to predict a higher risk of deterioration with $Spearman$ since the average absolute influence of these attributes is higher for the aggravation class. 

All these behaviors from our model are coherent with the clinical literature about Covid-19 \cite{zheng2020risk}. In contrast, with the $Kernel SHAP$ method, the near-zero average influences for some attributes are inconsistent with known risk factors.


\begin{figure}[h]
    \centering
    \includegraphics[width=\textwidth]{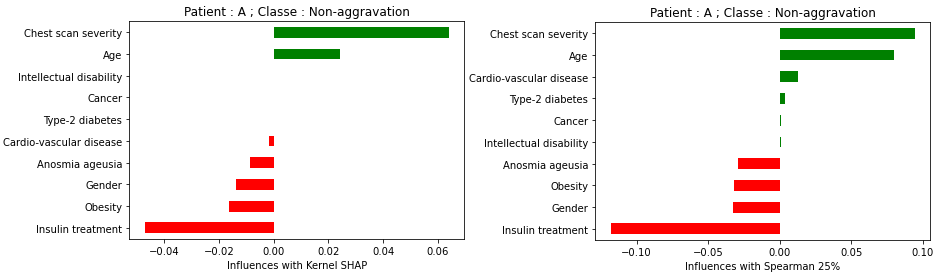}
    \caption{Influences of patient A with Kernal SHAP and Spearman 25\%}
    \label{fig:infs_instance_A}
\end{figure}

Another important point of the explanations is the fidelity and ease of understanding and interpreting them. Although very subjective, these parameters are essential to take into account in the medical field, since a lack of fidelity to the model and understanding of the explanations can lead to wrong decision making and consequences for the health of patients. To evaluate this, one instance of each class from the Covid-19 dataset was randomly drawn to describe and evaluate the explanation of the $Kernel SHAP$ and the $Spearman$ coalitional method. Figures \ref{fig:infs_instance_A} and \ref{fig:infs_instance_B} show the influence of each attribute for these patients, whose descriptions are given below.
Patient A is a 54-year-old obese man with no clinic sign of infection in his chest CT scan. He also has insulin treatment and signs of ageusia or anosmia. The two methods find that the value of Chest CT scan severity and age for this patient contributes the most to the prediction of non-aggravation while his gender, his symptoms of anosmia and ageusia, his obesity, and his insulin treatment goes against the prediction. The explanations allow us to understand that this patient has many risk factors and that the non-aggravation prediction comes mainly from the absence of severity of the chest CT scan and the patient's age.
However, for the \textit{Kernel SHAP} method, the absence of cardiovascular disease goes against non-aggravation prediction while it contributes to the prediction for the \textit{Spearman} method. This seems contrary to medical knowledge about Covid-19 \cite{zheng2020risk}, since cardiovascular disease is a risk factor. The absence of disease should therefore be in favour of a non-aggravation of the patient's state of health. 

Patient B is a 76-year-old man with type-2 diabetes, insulin treatment, ageusia anosmia. The severity of his chest CT scan is 4 out of 4 which is a critical value. For \textit{Kernel SHAP} method, the chest scan severity is way more important than other attributes in the prediction. For \textit{Spearman} method, even if the severity of the chest CT scan is significant, the presence of insulin treatment, the patient's gender and age are important. The absence of cancer, cardiovascular disease, intellectual disability, and obesity goes against the prediction, while there are no impact with \textit{Kernel SHAP} method. \textit{Spearman}'s explanations are slightly more contrasted than \textit{Kernel SHAP}'s ones.  

For this use case, the two methods are easy to understand as there are based on the same additive strategy. For both methods and both examples, influences approximate closely model predictions, and therefore have a high fidelity. However, this fidelity is only local, as methods only explain individual data instances. Moreover, and based on the clinical literature about Covid-19 \cite{zheng2020risk}, the explanations from \textit{Spearman} method seem more consistent for comorbidities. Finally, the complete dataset was computed in 51 seconds with the \textit{Spearman} method when it took more than 18 minutes for the \textit{Kernel SHAP} method, for similar results.

\begin{figure}[h]
    \centering
    \includegraphics[width=\textwidth]{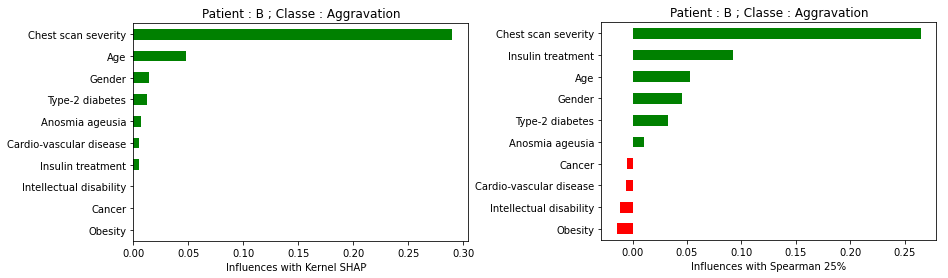}
    \caption{Influences of patient B with Kernal SHAP and Spearman 25\%}
    \label{fig:infs_instance_B}
\end{figure}

\section{Source code}

The full implementation of our proposals (including all the methods proposed in Section \ref{sec:new_methods} as well as the \textit{K-complete} and \textit{Complete} methods) is available here: \url{https://github.com/kaduceo/coalitional_explanation_methods}.\\
The source code will evolve considering future works.

\section{Conclusion and perspectives}
\label{sec:conclusion}

This paper explored several approximations of the additive explanation method based on Shapley values, named \textit{complete} method in the paper since this method is computationally exponential with respect to attribute number, thus intractable in most practical cases.
We compared existing approximation methods such as \textit{SHAP} or \textit{k-depth} that are both limited in their inclusion of attribute interdependence which has a clear impact on their performances either in their accuracy, with respect to the \textit{complete} method, or computation time.
In order to  take into account the interaction between attributes more efficiently, we developed several methods, called \textit{coalitional} methods, based on smarter grouping attribute procedures that only retain relevant groups of attributes, thus lowering complexity and computation time while maintaining an acceptable precision. 
Tests were conducted on 243 datasets with a different number of attributes and instances. \textit{Coalitional} methods show the most promising results, especially those based on \textit{PCA}, \textit{Spearman} and \textit{Reverse VIF}. These new methods notably outperform existing ones, whether \textit{k-depth} or \textit{SHAP}-based, for more complex datasets where attribute interdependence is more likely to be present. 
Although our results open up encouraging perspectives of practical application of individual prediction interpretability, either in terms of accuracy or in computation time, the main problem is that computing the \textit{complete} influences, which is the comparison baseline for our study, becomes near impossible with larger attribute numbers. Thus, it is very difficult to monitor the performance of our different methods with this baseline.
A possible way to address this problem could be first to run a global attribute importance study for large datasets using methods such as \textit{Permutation Importance} that is model agnostic or \textit{Gini Importance} for tree-based models. Then use this information to compute influences only for the most important attributes during the individual explanation generation.

A longer-term perspective is also to take into account the context where the predictive analysis is conducted. Indeed, the explanations provided for particular instances cannot be totally satisfactory for any users in any situations. The degree of user expertise in the data seems very important to consider: an expert user will certainly be more interested in very precise explanations than a novice one.
Moreover, the analysis process may have an impact on the type of explanations to consider. The explanations should not be analysed in the same way depending on whether the analysis is carried out in an exploratory or confirmatory manner.

%In this paper, we proposed a comparative study between several attribute grouping methods (inspired by the feature selection field) in an objective of individual prediction explanation.
%Our tests, conducted with 324 real datasets, show that  \textit{RevVIF}, \textit{PCA} and \textit{Model} methods are all of interest. \textit{RevVIF} is preferable for datasets with few attributes, while \textit{PCA} and \textit{Model} should fare better for a large set of attributes. Then, a new interesting perspective would be to study the evolution of computation times with larger datasets. The main problem here is it becomes impossible to compute the \textit{Complete influence} for large datasets. Thus, it is impossible to monitor the performance of our different methods with this baseline.
%To address this problem, a possible way could be to run a general attribute importance study for large datasets, first, and use this information to calculate the influence of the most important attributes during the individual explanation generation.

\section{Acknowledgement}

The use-case dataset was acquired in collaboration with the Centre Hospitalier Intercommunal de Créteil. Therefore, we greatly acknowledge the managers and physicians involved in this project.

\bibliographystyle{unsrt}  
\bibliography{biblio}

\begin{thebibliography}{10}

\bibitem{adadi2018peeking}
Amina Adadi and Mohammed Berrada.
\newblock Peeking inside the black-box: A survey on explainable artificial
  intelligence (xai).
\newblock {\em IEEE Access}, 6:52138--52160, 2018.

\bibitem{carvalho2019machine}
Diogo~V Carvalho, Eduardo~M Pereira, and Jaime~S Cardoso.
\newblock Machine learning interpretability: A survey on methods and metrics.
\newblock {\em Electronics}, 8(8):832, 2019.

\bibitem{strumbelj_efficient_2010}
Erik Strumbelj and Igor Kononenko.
\newblock An {Efficient} {Explanation} of {Individual} {Classifications}
  {Using} {Game} {Theory}.
\newblock {\em J. Mach. Learn. Res.}, 11:1--18, March 2010.
\newblock Publisher: JMLR.org.

\bibitem{casalicchio2018visualizing}
Giuseppe Casalicchio, Christoph Molnar, and Bernd Bischl.
\newblock Visualizing the feature importance for black box models.
\newblock In {\em Joint European Conference on Machine Learning and Knowledge
  Discovery in Databases}, pages 655--670. Springer, 2018.

\bibitem{lundberg2017consistent}
Scott~M Lundberg and Su-In Lee.
\newblock Consistent feature attribution for tree ensembles.
\newblock {\em arXiv preprint arXiv:1706.06060}, 2017.

\bibitem{vstrumbelj2008towards}
Erik {\v{S}}trumbelj and Igor Kononenko.
\newblock Towards a model independent method for explaining classification for
  individual instances.
\newblock In {\em International Conference on Data Warehousing and Knowledge
  Discovery}, pages 273--282. Springer, 2008.

\bibitem{broeck2020tractability}
Guy~Van den Broeck, Anton Lykov, Maximilian Schleich, and Dan Suciu.
\newblock On the tractability of shap explanations, 2020.

\bibitem{10.1007/978-3-030-38919-2_26}
Gabriel Ferrettini, Julien Aligon, and Chantal Soul{\'e}-Dupuy.
\newblock Explaining single predictions: A faster method.
\newblock In Alexander Chatzigeorgiou, Riccardo Dondi, Herodotos Herodotou,
  Christos Kapoutsis, Yannis Manolopoulos, George~A. Papadopoulos, and Florian
  Sikora, editors, {\em SOFSEM 2020: Theory and Practice of Computer Science},
  pages 313--324, Cham, 2020. Springer International Publishing.

\bibitem{DBLP:conf/adbis/FerrettiniAS20}
Gabriel Ferrettini, Julien Aligon, and Chantal Soul{\'{e}}{-}Dupuy.
\newblock Improving on coalitional prediction explanation.
\newblock In J{\'{e}}r{\^{o}}me Darmont, Boris Novikov, and Robert Wrembel,
  editors, {\em Advances in Databases and Information Systems - 24th European
  Conference, {ADBIS} 2020, Lyon, France, August 25-27, 2020, Proceedings},
  volume 12245 of {\em Lecture Notes in Computer Science}, pages 122--135.
  Springer, 2020.

\bibitem{lundberg_unified_2017}
Scott~M Lundberg and Su-In Lee.
\newblock A {Unified} {Approach} to {Interpreting} {Model} {Predictions}.
\newblock In I.~Guyon, U.~V. Luxburg, S.~Bengio, H.~Wallach, R.~Fergus,
  S.~Vishwanathan, and R.~Garnett, editors, {\em Advances in {Neural}
  {Information} {Processing} {Systems} 30}, pages 4765--4774. Curran
  Associates, Inc., 2017.

\bibitem{altmann_permutation_2010}
André Altmann, Laura Toloşi, Oliver Sander, and Thomas Lengauer.
\newblock Permutation importance: a corrected feature importance measure.
\newblock {\em Bioinformatics}, 26(10):1340--1347, 2010.

\bibitem{kira1992practical}
Kenji Kira and Larry~A Rendell.
\newblock A practical approach to feature selection.
\newblock In {\em Machine Learning Proceedings 1992}, pages 249--256. Elsevier,
  1992.

\bibitem{henelius2017interpreting}
Andreas Henelius, Kai Puolam{\"a}ki, and Antti Ukkonen.
\newblock Interpreting classifiers through attribute interactions in datasets.
\newblock {\em arXiv preprint arXiv:1707.07576}, 2017.

\bibitem{wachter2017counterfactual}
Sandra Wachter, Brent Mittelstadt, and Chris Russell.
\newblock Counterfactual explanations without opening the black box: Automated
  decisions and the gdpr.
\newblock {\em Harv. JL \& Tech.}, 31:841, 2017.

\bibitem{wexler2019if}
James Wexler, Mahima Pushkarna, Tolga Bolukbasi, Martin Wattenberg, Fernanda
  Vi{\'e}gas, and Jimbo Wilson.
\newblock The what-if tool: Interactive probing of machine learning models.
\newblock {\em IEEE transactions on visualization and computer graphics},
  26(1):56--65, 2019.

\bibitem{Strumbelj:2010:EEI:1756006.1756007}
E.~Strumbelj and I.~Kononenko.
\newblock Explaining prediction models and individual predictions with feature
  contributions.
\newblock {\em Knowledge and Information Systems}, 41:647--665, 2013.

\bibitem{datta_algorithmic_2016}
A.~{Datta}, S.~{Sen}, and Y.~{Zick}.
\newblock Algorithmic transparency via quantitative input influence: Theory and
  experiments with learning systems.
\newblock In {\em 2016 IEEE Symposium on Security and Privacy (SP)}, pages
  598--617, May 2016.

\bibitem{ribeiro2016should}
Marco~Tulio Ribeiro, Sameer Singh, and Carlos Guestrin.
\newblock " why should i trust you?" explaining the predictions of any
  classifier.
\newblock In {\em Proceedings of the 22nd ACM SIGKDD international conference
  on knowledge discovery and data mining}, pages 1135--1144, 2016.

\bibitem{elshawi2020interpretability}
Radwa ElShawi, Youssef Sherif, Mouaz Al-Mallah, and Sherif Sakr.
\newblock Interpretability in healthcare: A comparative study of local machine
  learning interpretability techniques.
\newblock {\em Computational Intelligence}, 2020.

\bibitem{Bolon2013}
Verónica Bolón-Canedo, Noelia Sánchez-Maroño, and Amparo Alonso-Betanzos.
\newblock A review of feature selection methods on synthetic data.
\newblock {\em Knowledge and Information Systems}, 34(3):483--519, 2013.

\bibitem{Yu2004}
Lei Yu and Huan Liu.
\newblock Efficient feature selection via analysis of relevance and redundancy.
\newblock {\em J. Mach. Learn. Res.}, 5:1205–1224, December 2004.

\bibitem{Rakotomamonjy2003}
Alain Rakotomamonjy.
\newblock Variable selection using svm based criteria.
\newblock {\em J. Mach. Learn. Res.}, 3(null):1357–1370, March 2003.

\bibitem{Mejia2006}
M~Mejía-Lavalle, E~Sucar, and G~Arroyo.
\newblock Variable selection using svm based criteria.
\newblock In {\em International workshop on feature selection for data mining},
  page 131–1350, 2006.

\bibitem{hall1999correlationbased}
Mark~A. Hall.
\newblock {\em Correlation-based Feature Selection for Machine Learning}.
\newblock PhD thesis, 1999.

\bibitem{shapley1953}
L.~S. Shapley.
\newblock A value for n-person games.
\newblock {\em Contributions to the Theory of Games}, (28):307--317, 1953.

\bibitem{10.1093/nar/gkaa219}
Simon Eitzinger, Amina Asif, Kyle~E Watters, Anthony~T Iavarone, Gavin~J Knott,
  Jennifer~A Doudna, and Fayyaz ul Amir Afsar Minhas.
\newblock {Machine learning predicts new anti-CRISPR proteins}.
\newblock {\em Nucleic Acids Research}, 48(9):4698--4708, 04 2020.

\bibitem{9233366}
E.~{Tjoa} and C.~{Guan}.
\newblock A survey on explainable artificial intelligence (xai): Toward medical
  xai.
\newblock {\em IEEE Transactions on Neural Networks and Learning Systems},
  pages 1--21, 2020.

\bibitem{Bibault2020}
Jean-Emmanuel Bibault, Daniel Chang, and Lei Xing.
\newblock Development and validation of a model to predict survival in
  colorectal cancer using a gradient-boosted machine.
\newblock {\em Gut}, 09 2020.

\bibitem{lauritsen2020explainable}
Simon~Meyer Lauritsen, Mads Kristensen, Mathias~Vassard Olsen, Morten~Skaarup
  Larsen, Katrine~Meyer Lauritsen, Marianne~Johansson J{\o}rgensen, Jeppe
  Lange, and Bo~Thiesson.
\newblock Explainable artificial intelligence model to predict acute critical
  illness from electronic health records.
\newblock {\em Nature communications}, 11(1):1--11, 2020.

\bibitem{Ribeiro2016}
Marco~Tulio Ribeiro, Sameer Singh, and Carlos Guestrin.
\newblock "why should i trust you?": Explaining the predictions of any
  classifier.
\newblock In {\em Proceedings of the 22Nd ACM SIGKDD International Conference
  on Knowledge Discovery and Data Mining}, KDD '16, pages 1135--1144, New York,
  NY, USA, 2016. ACM.

\bibitem{shrikumar_learning_2017}
Avanti Shrikumar, Peyton Greenside, and Anshul Kundaje.
\newblock Learning {Important} {Features} {Through} {Propagating} {Activation}
  {Differences}.
\newblock In {\em Proceedings of the 34th {International} {Conference} on
  {Machine} {Learning} - {Volume} 70}, {ICML}'17, pages 3145--3153, 2017.
\newblock event-place: Sydney, NSW, Australia.

\bibitem{10.1371/journal.pone.0130140}
Sebastian Bach, Alexander Binder, Grégoire Montavon, Frederick Klauschen,
  Klaus-Robert Müller, and Wojciech Samek.
\newblock On pixel-wise explanations for non-linear classifier decisions by
  layer-wise relevance propagation.
\newblock {\em PLOS ONE}, 10(7):1--46, 07 2015.

\bibitem{Lipovetsk2001}
Stan Lipovetsky and Michael Conklin.
\newblock Analysis of regression in game theory approach.
\newblock {\em Applied Stochastic Models in Business and Industry}, 17:319 --
  330, 10 2001.

\bibitem{Henelius1175294}
Andreas Henelius, Kai Puolamaki, Henrik Bostr{\"o}m, Lars Asker, and Panagiotis
  Papapetrou.
\newblock A peek into the black box : exploring classifiers by randomization.
\newblock {\em Data mining and knowledge discovery}, 28(5-6):1503--1529, 2014.
\newblock QC 20180119.

\bibitem{makki2019efficient}
Sara Makki.
\newblock {\em An Efficient Classification Model for Analyzing Skewed Data to
  Detect Frauds in the Financial Sector}.
\newblock PhD thesis, Universit{\'e} de Lyon; Universit{\'e} libanaise, 2019.

\bibitem{Vanschoren2013}
Joaquin Vanschoren, Jan~N. van Rijn, Bernd Bischl, and Luis Torgo.
\newblock Openml: Networked science in machine learning.
\newblock {\em SIGKDD Explorations}, 15(2):49--60, 2013.

\bibitem{robnik-sikonja_perturbation-based_2018}
Marko Robnik-Sikonja and Marko Bohanec.
\newblock Perturbation-{Based} {Explanations} of {Prediction} {Models}.
\newblock In {\em Human and {Machine} {Learning}}, pages 159--175. June 2018.

\bibitem{francone2020chest}
Marco Francone, Franco Iafrate, Giorgio~Maria Masci, Simona Coco, Francesco
  Cilia, Lucia Manganaro, Valeria Panebianco, Chiara Andreoli, Maria~Chiara
  Colaiacomo, Maria~Antonella Zingaropoli, et~al.
\newblock Chest ct score in covid-19 patients: correlation with disease
  severity and short-term prognosis.
\newblock {\em European radiology}, 30(12):6808--6817, 2020.

\bibitem{zheng2020risk}
Zhaohai Zheng, Fang Peng, Buyun Xu, Jingjing Zhao, Huahua Liu, Jiahao Peng,
  Qingsong Li, Chongfu Jiang, Yan Zhou, Shuqing Liu, et~al.
\newblock Risk factors of critical \& mortal covid-19 cases: A systematic
  literature review and meta-analysis.
\newblock {\em Journal of Infection}, 2020.

\end{thebibliography}

\end{document}